\documentclass[]{fairmeta}

\usepackage{hyperref}
\usepackage{url}
\usepackage{xcolor}
\usepackage{graphicx}
\usepackage{booktabs}
\usepackage{amssymb}
\usepackage{pifont}
\usepackage{makecell}
\usepackage{multirow}
\usepackage{colortbl}
\usepackage{algorithm}
\usepackage{algpseudocode}
\usepackage{wrapfig}
\newcommand{\greencheck}{\textcolor{green!55!black}{\ding{52}}}
\newcommand{\redcross}{\textcolor{red!75!black}{\ding{55}}}
\usepackage{float}
\usepackage{color}
\usepackage{soul}
\definecolor{lightred}{rgb}{1, 0.8, 0.8}
\definecolor{lightblue}{rgb}{0.8, 0.9, 1}
\usepackage[most]{tcolorbox}
\usepackage{listings}
\usepackage{float}

\tcbset{
  aibox/.style={
    width=390pt,
    top=10pt,
    colback=white,
    colframe=black,
    colbacktitle=black,
    enhanced,
    center,
    attach boxed title to top left={yshift=-0.1in,xshift=0.15in},
    boxed title style={boxrule=0pt,colframe=white,},
  }
}
\newtcolorbox{AIbox}[2][]{aibox,title=#2,#1}

\title{MemLife: Curating and Reasoning over Long-Term Egocentric Video Memories}

\author[1,2,\dag,*]{Guangzhi Xiong}
\author[1,\dag]{Xinyuan Zhang}
\author[1]{Xiao Yang}
\author[1]{Hyokun Yun}
\author[1]{Kai Zhang}
\author[1]{Shiun-Zu Kuo}
\author[1,3,*]{Hyeonjeong Ha}
\author[1]{Xilun Chen}
\author[1]{Kai Sun}
\author[1]{Lucas Liang}
\author[1]{Guangqiang Dong}
\author[1]{Ejaz Ahmed}
\author[1]{Ahmed A Aly}
\author[1]{Anuj Kumar}
\author[1]{Raffay Hamid}
\author[2,\dag]{Aidong Zhang}
\author[1,\dag]{Xin Luna Dong}

\affiliation[1]{Meta Reality Labs}
\affiliation[2]{University of Virginia}
\affiliation[3]{University of Illinois Urbana-Champaign}

\contribution[*]{Work done at Meta}

\abstract{Long-term egocentric video enables personalized AI assistants to reason about daily life.
However, as video histories grow to hundreds of hours spanning months or years, reprocessing raw clips for every query becomes computationally prohibitive.
Memory systems offer a scalable alternative by compacting videos into text representations, but often fail on practical benchmarks: either the memory does not preserve key evidence, or the retriever fails to locate relevant entries due to retrieval competition in growing search spaces.
To address these challenges, we introduce \textsc{MemLife}, a multimodal memory system that constructs entity-grounded, first-person text episodes and retrieves them via a time-indexed agentic reader.
Without training or query-time video access, \textsc{MemLife} improves over the strongest training-free baseline by 4.6--12.0\% across four long-horizon benchmarks.
To further improve memory quality, we propose \textsc{MemOpt}, a reinforcement learning framework that 
optimizes the memory writer to produce faithful, informative, and retrievable memories.
\textsc{MemOpt} consistently improves \textsc{MemLife} by 2.7--5.0\% across different video and question distributions, with gains that generalize across writer and reader backbones and memory systems.}

\correspondence{\textsuperscript{\dag}\email{\{guangzhi,aidong\}@virginia.edu}, \email{\{dylanz426,lunadong\}@meta.com}}
\metadata[Code]{\url{https://github.com/facebookresearch/MemLife}}

\begin{document}

\maketitle

\section{Introduction}
If an AI assistant could record a user's life experiences as egocentric videos, captured by wearable devices such as smart glasses and GoPro cameras~\citep{grauman2022ego4d}, \textit{could it then answer any question the user asks about their past?} At first glance, this is a Retrieval-Augmented Generation (RAG) problem over egocentric videos~\citep{yang2025egolife,alam2026supermemory}. However, question answering (QA) over such \textit{dense} memories is substantially harder. On the data side, videos accumulate over time to \textit{prohibitive volumes}, and scenes and events \textit{recur with subtle differences}. On the system side, the \textit{latency budget} for sifting through long, similar memories is tight, and the \textit{context window} is too small to hold many visual frames. It is therefore crucial to \textit{compact raw videos into efficient representations} that can serve as the primary source of evidence for downstream tasks. Natural-language text stands out as an appealing choice: orders of magnitude more compact than visual frames, natively consumable by Large Language Models (LLMs), and interpretable to users.

\begin{figure}[h!]
    \centering
    \includegraphics[width=1.0\linewidth]{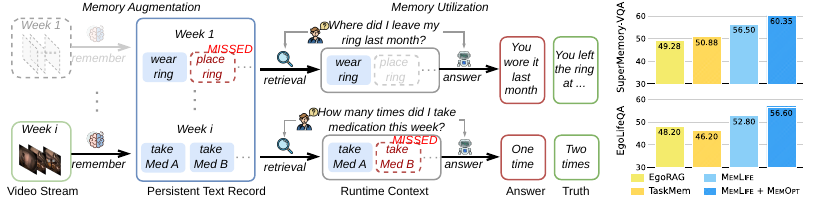}
    \caption{Failure modes of long-term video memory systems. Dotted objects denote lost information (e.g., Week-1 video is deleted). Systems fail when relevant evidence is omitted during memory writing or missed during retrieval. Our proposed solutions outperform prior state-of-the-art systems.}
    \label{fig:motivation}
\end{figure}

We follow the line of work that converts multimodal video recordings into textual descriptions~\citep{islam2024videorecap,yang2025egolife}, either to facilitate retrieval of key evidence~\citep{luo2025videorag,ren2025videorag} or to directly support downstream generation~\citep{yeo2026worldmm,yin2026videoarm}. However, deciding what to write poses a fundamental tradeoff (Figure~\ref{fig:motivation}): \textit{aggressive} compression may discard information that future questions ask about, whereas \textit{conservative} compression retains trivial details that dilute retrieval and inflate QA-time processing. Worse, writers may \textit{hallucinate} content unsupported by the video, planting false evidence. In this paper, we address a critical question for QA over long-term memory: how can a system strike the right balance to \textit{remember only what is worth remembering, faithfully,} and \textit{in a form that is easy to recall}?

We answer this question in two steps: we first design a strong writer by hand, then learn a better one.
Our first contribution, \textsc{MemLife}, is an agentic memory system whose writer follows two principles. 
First, it \textit{anchors memories in time and grounds entities across modalities}, aligning spoken references with the people and objects observed in each episode.
Second, it \textit{narrates in the first person}, matching how users phrase questions about their own lives (e.g., ``Where did I put my passport?'') and thereby narrowing the query-memory gap in agentic retrieval. 
To exploit these memories, \textsc{MemLife}'s reader combines agentic semantic search with time-scoped memory fetching, and presents retrieved episodes in chronological order for reasoning over long histories.
In the presence of source videos, the agentic reader selectively invokes video-retrieval tools to sample raw video frames whenever visual details are required.
Without writer training or query-time video access, \textsc{MemLife} improves over the strongest training-free baseline by up to 12.0\% in accuracy.

Our second insight is that \textit{learning what to write} is, by itself, a powerful lever for long-term memory QA. We therefore take a bold step: rather than applying Reinforcement Learning (RL) to improve final answer quality~\citep{guan2026videostreamingthinking,yan2025memoryr1,wang2025mema,li2026ember}, we apply RL \textit{only} to memory writing, and examine whether this alone improves memory QA. We propose \textsc{MemOpt}, a learning framework that rewards \textit{Faithful}, \textit{Informative}, and \textit{Retrievable} Memories, a reward system we call \textit{FIRM}. Faithfulness penalizes hallucinated memories unsupported by the source video; informativeness encourages the writer to preserve details critical to answering future questions; and retrievability favors concise descriptions that enable easy and precise retrieval. Although \textsc{MemOpt} trains only the writer and does not rely on supervision from stronger models~\citep{long2026seeing,zou2026task}, it further improves \textsc{MemLife} by 2.7--5.0\%.

Our paper makes the following contributions.
\begin{itemize}
    \item We introduce \textsc{MemLife}, a multimodal memory system that compresses egocentric videos into time- and entity-anchored, first-person episodes, and reasons over them with a versatile agentic reader.
    \item We propose \textsc{MemOpt}, a learning framework whose FIRM reward optimizes the memory writer with multi-granular feedback on faithfulness, informativeness, and retrievability, to address the write-time failure modes we identified. 
    \item We show that \textsc{MemOpt} combined with \textsc{MemLife} outperforms the strongest prior baseline by 8.4--17.0\%, while reducing the memory size by up to 31$\times$ (Appendix~\ref{app:efficiency}). The trained writer transfers well: it consistently improves accuracy when plugged into other memory systems (e.g., EgoRAG) and backbones, and on out-of-domain videos and question types.
\end{itemize}

\section{Related Work}
\label{sec:related_work}

\noindent\textbf{Writing memory over extended video horizons.}
Memory-augmented video paradigms differ in how they convert continuous streams into persistent representations.
Retrieval-oriented systems encode local clips as flat text descriptions or visual-text indexes~\citep{fan2024videoagent, islam2024videorecap, luo2025videorag}.
To control memory growth, streaming architectures maintain fixed-budget recurrent buffers that continuously compress past frames~\citep{guan2026videostreamingthinking, he2024malmm, qian2024videostreaming, song2024moviechat, jin2025videomem}, though 
they often struggle to preserve distant or fine-grained details.
Hierarchical frameworks summarize video streams across temporal tiers~\citep{islam2024videorecap, yang2025egolife}, while structured memory agents organize representations around entities, events, or scene graphs~\citep{goletto2024amego, long2026seeing, ren2025videorag, yeo2026worldmm, yin2026videoarm}.
However, high-level abstractions suffer from information loss, and graph maintenance becomes computationally prohibitive.
Furthermore, prior memory construction relies on heuristic prompt engineering rather than optimizing memory generation via task feedback.

\noindent\textbf{Reading memory across multi-session histories.}
Personal memory systems rely on readers to locate evidence scattered across extended temporal histories.
Passive retrieval-based readers execute single-shot semantic or temporal queries over text indices~\citep{lewis2020rag, luo2025videorag, yang2025egolife, ren2025videorag}.
Recent egocentric architectures introduce specialized access mechanisms—such as Memory Pointer Prompting~\citep{ye2025mmego} or multi-turn reasoning loops that reformulate queries, fetch time intervals, and inspect visual frames~\citep{yeo2026worldmm, yin2026videoarm, gao2023assistgpt, wang2025videoagent, tian2026egor1, zhang2025deepvideo}.
While effective during inference~\citep{chandrasegaran2024hourvideo, wang2026egomemreason}, using multi-turn readers during writer post-training conflates memory quality with reader execution noise.
Because task accuracy depends on sampled tool calls, end-to-end task rewards provide a noisy, computationally prohibitive supervision signal for writer optimization.

\noindent\textbf{Training video memory writers.}
To move beyond heuristic prompt design, recent paradigms train memory writers using learned policies.
One direction relies on supervised fine-tuning or imitation learning, distilling memory construction from proprietary demonstrations or QA instructions (\emph{e.g.}, EgoButler~\citep{yang2025egolife}, M3-Agent~\citep{long2026seeing}).
Another direction employs reinforcement learning, optimizing policies like TaskMem~\citep{zou2026task} and VST~\citep{guan2026videostreamingthinking} against downstream task accuracy or QA preferences.
However, static distillation restricts writer adaptability, while training purely on end-to-end task rewards introduces severe execution noise from reader reasoning.
In contrast, \textsc{MemOpt} provides supervision from verified evidence, source video, and fixed reader actions, decoupling writer post-training from reader execution noise.
\section{Methodology}
\label{sec:methodology}

\subsection{Problem Definition and Solution Overview}
We begin by defining the {\em Video Memory QA} problem. Consider a stream of video (optionally egocentric) $\mathcal{V}=(c_1,\ldots,c_T)$ of $T$ segments. Each segment can be represented as a triplet $c_t=(F_t,S_t,\tau_t)$, where $F_t$ denotes the sampled visual frame, $S_t$ denotes the aligned audio transcriptions, and $\tau_t=[s_t,e_t]$ denotes the starting and ending time. Memory QA takes a question $q$ arriving at time $\tau_q$, and provides the answer based on the prior memory fragments:
\begin{equation}
    \mathcal{V}_q=\{c_t\in\mathcal{V}:e_t\leq\tau_q\}.
    \label{eq:memlife}
\end{equation}

Our first solution \textsc{MemLife} converts a video stream into persistent episodic memory and uses an agentic reader to retrieve and reason over relevant entries. Formally, \textsc{MemLife} employs a writer $W_\theta$ with parameters $\theta$; the writer generates a textual description $d_t$ for each memory segment $m_t$:
\begin{equation}
    d_t\sim W_\theta(\cdot\mid F_t,S_t),
    \qquad
    m_t=(d_t,\tau_t).
    \label{eq:memlife_writer}
\end{equation}

Thus, the memory repository stores $\mathcal{M}(\mathcal{V})=\{m_t\}_{t=1}^{T}$. For a question $q$, question answering uses the available memory $\mathcal{M}_q$, optionally with the corresponding source-video history $\mathcal{V}_q$, where
\begin{equation}
    \mathcal{M}_{q}=\{m_t\in\mathcal{M}(\mathcal{V}):e_t\leq\tau_q\}.
    \label{eq:memlife_reader}
\end{equation}

\begin{figure}[h!]
    \begin{center}
    \includegraphics[width=1\linewidth]{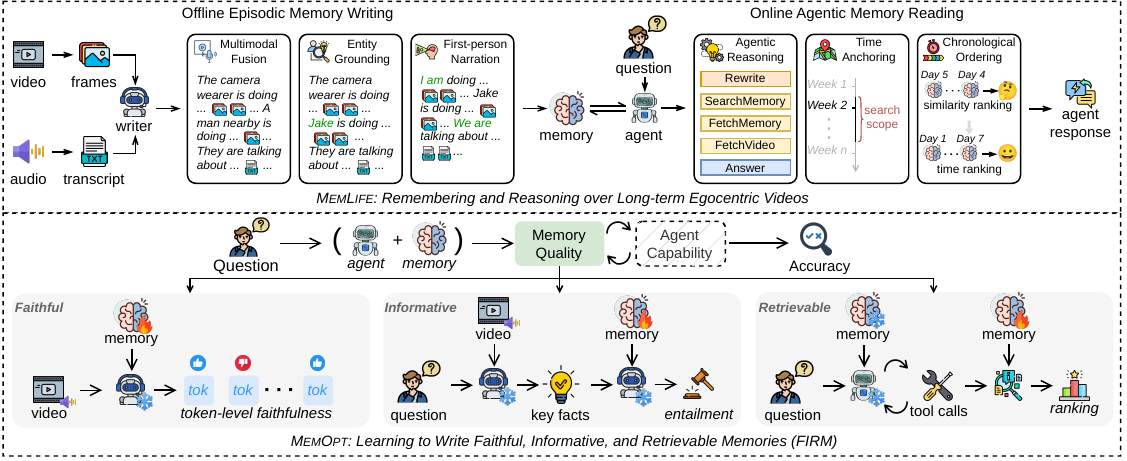}
    \end{center}
    \caption{Overview of \textsc{MemLife}'s episodic writer and time-indexed agentic reader, together with \textsc{MemOpt}'s decomposed writer supervision. Direct source-video access is optional for the reader.}
    \label{fig:overview}
\end{figure}

Our second solution, \textsc{MemOpt}, is a training framework that improves the memory writer by optimizing its parameters $\theta$. Figure \ref{fig:overview} gives the overview of the two solutions.

\subsection{\textsc{MemLife} System for Memory Writing and Reading}

{\bf Writer:} \textsc{MemLife} separates question-independent memory construction from question-dependent memory access. 
Because future questions are unknown during writing, each entry must preserve information across modalities and express it in a form that supports later retrieval. The \textsc{MemLife} writer applies three designs for this purpose.

\noindent\textit{Multimodal fusion.} Information needed by future questions may appear in either the visual stream or speech. The writer therefore jointly interprets the frames and transcript such that evidence from both modalities can be preserved in one memory entry.

\noindent\textit{Entity grounding.} A transcript may mention an entity by its name, which provides an important cue for future QA. The writer aligns the speech with the memory segment and uses the identified name to refer to the visual referents in the description. 

\noindent\textit{First-person narration.} Memory questions for egocentric videos naturally refer to the user as ``I.'' The writer therefore adopts the same first-person perspective, making its descriptions easier to match future egocentric questions.

\textsc{MemLife} generates a description and its embedding for every clip independently, such that it avoids sequential dependencies and error propagation, and keeps total computation and storage linear in the recorded history. We also explored conditioning the writer on textual or multimodal context from the preceding segments, but neither variant improves aggregate accuracy (Appendix~\ref{app:predecessor_context}).

{\bf Reader:} The \textsc{MemLife} reader is an agentic system that takes a question $q$ and its timestamp $\tau_q$ as input and operates over multiple rounds by selecting actions from the action space $\cal A$:
\begin{equation}
\begin{split}
     \mathcal{A}=\{&\textsc{Rewrite}(u, I| q, \tau_q),\ \textsc{SearchMemory}(\bar M|u,I,k),\ \textsc{FetchMemory}(\bar M|I),\\ 
     & \textsc{FetchVideo}(\bar V|I,f),\ \textsc{Answer}(a|q, \bar M, \bar V)\}.
\end{split}
\label{eq:memlife_actions}
\end{equation}

With \textsc{Rewrite}, the agent reasons over the current information, and transforms the input $(q, \tau_q)$ into a targeted search query $u$ and/or a time interval $I$ over the history. For a search query $u$, \textsc{SearchMemory} conducts the similarity search and returns $k$ relevant entries $\bar M$ from the stored memories within interval $I$. With only interval $I$, the agent can call either \textsc{FetchMemory}, which returns all text memory entries $\bar M$ within $I$, or \textsc{FetchVideo}, which returns raw multimodal fragments $\bar V$ and $f$ sampled frames within $I$.
Finally, \textsc{Answer} generates the answer $a$ based on retrieval results.

The \textsc{FetchVideo} tool is disabled when source videos are unavailable at reading time. For video-available settings (\textsc{MemLife}-V), we store low-resolution redacted videos due to storage and privacy concerns, and sample limited frames to optimize computation.

\subsection{\textsc{MemOpt} Framework for Optimizing Memory Writer}

\textsc{MemOpt} updates the writer parameters $\theta$ through supervision, while keeping the agentic reader fixed. We next present our FIRM reward model, the major recipe to improve memory writing. 

\textbf{Theoretical foundation.}
The goal of \textsc{MemOpt} is to teach the writer what is worth remembering and which form is easy to recall. We next show the theoretical quantification.

Let $Q$ and $A$ denote a random question and its answer, while $\cal V$, $\cal M$, and $C_Q$ denote the available video history, its corresponding memory, and the context retrieved by the reader for $Q$.
To isolate the quality of the written memory, we restrict the reader's evidence source to $\cal M$, excluding direct access to $\cal V$ that could otherwise bypass the memory.
Because the writer rewrites $\cal V$ into $\cal M$, and the reader constructs $C_Q$ only from $(Q, {\cal M})$, their joint distribution factorizes as
\begin{equation}
    p(Q,A,{\cal V},{\cal M},C_Q)
    =p(Q,A,{\cal V}) \cdot
    p_\theta({\cal M}\mid {\cal V}) \cdot
    p(C_Q\mid Q, {\cal M}).
    \label{eq:memory_factorization}
\end{equation}
Let $H(\mid)$ denote conditional entropy and $I(\mid)$ conditional mutual information. Under this factorization, the additional uncertainty about the answer $A$ when using the retrieved context $C_Q$ instead of the video history $\cal V$ decomposes exactly as

\begin{equation}
\begin{aligned}
    \underbrace{H(A\mid Q, C_Q)-H(A\mid Q, {\cal V})}_{\text{total information loss}}
    ={}&\underbrace{I(A;{\cal V}\mid Q, {\cal M})}_{\Delta_{\mathrm{inf}}\text{: memory writing loss}}
    +\underbrace{I(A;{\cal M}\mid Q, C_Q)}_{\Delta_{\mathrm{ret}}\text{: memory recall loss}}.
\end{aligned}
    \label{eq:memory_information_decomposition}
\end{equation}

The equation shows that answer-relevant information can be lost when the writer maps $\cal V$ to $\cal M$ ($\Delta_{\mathrm{inf}}$) and when the reader retrieves $C_Q$ from $\cal M$ ($\Delta_{\mathrm{ret}}$). These gaps motivate the informativeness and retrievability rewards, but do not measure whether $\cal M$ is grounded in $\cal V$. Unsupported memory can distort the final answer, while final-answer error also reflects answerer reasoning. We therefore assess faithfulness directly against the source video. Appendix~\ref{app:memory_decomposition} provides the complete derivation.

\textbf{Faithfulness.}
Faithfulness asks whether every claim in a candidate is supported by its source segment.
Because unsupported content may occupy only a few tokens, the feedback must also identify where it occurs. For candidate $y_i=(y_{i,1},\ldots,y_{i,L_i})$ with $L_i$ generated tokens under examination, we prompt the same frozen model to check the generated memory against its source segment by reproducing supported content exactly and minimally correcting unsupported spans, thereby localizing the grounding feedback. We denote by $p_{\mathrm{faith}}(w\mid c,y_i,y_{i,<t})$ the probability that the evaluator generates a possible next token $w$ and compute the faithfulness reward as
\begin{equation}
    R_{\mathrm{faith},i,t}
    =1-\left[
        \max_w p_{\mathrm{faith}}(w\mid c,y_i,y_{i,<t})
        -p_{\mathrm{faith}}(y_{i,t}\mid c,y_i,y_{i,<t})
    \right].
    \label{eq:reward_faithfulness}
\end{equation}
The reward lies in $[0,1]$. It equals 1 when the candidate token is the evaluator's most probable continuation and decreases when the evaluator favors a correction. Because each token is scored under its original prefix, the penalty concentrates on the first token of an unsupported span. Since later tokens are conditioned on this token, lowering its probability discourages the entire span. Human inspection confirms that most flagged spans are indeed unsupported (Appendix~\ref{app:rev_human_eval}).

\textbf{Informativeness.}
Informativeness asks whether the candidate itself preserves the answer-relevant evidence supplied by its source segment, independent of reader behavior.
We represent the required evidence as a source-grounded key fact and test whether the candidate entails it.
A frozen copy of the default writer model serves as both the key-fact \textit{extractor} and entailment \textit{judge}.
The question and answer are used only to compute the training reward, while the writer itself never receives them during training or deployment. 

Formally, let $c=c_t$ be a segment, $y$ be a candidate memory generated by the writer, $\mathcal{Q}(c)$ denote the set of questions for which $c$ provides verified evidence, and assume each question $q \in \mathcal{Q}(c)$ is paired with a correct answer $a_q$. For each $(q,a_q)$, the \textit{extractor} identifies the observed fact $k_{c,q}$ that supports the answer. Let $P_{\mathrm{ent}}(y\Rightarrow k_{c,q})$ denote the \textit{judge}'s estimated probability that $y$ entails this fact. We average this probability across the relevant questions,
\begin{equation}
    R_{\mathrm{inf}}(c,y)
    =\frac{1}{|\mathcal{Q}(c)|}
    \sum_{q\in\mathcal{Q}(c)}
    P_{\mathrm{ent}}(y\Rightarrow k_{c,q}).
    \label{eq:reward_informativeness}
\end{equation}

\textbf{Retrievability.}
Retrievability asks whether the relevant memory can be discovered at QA time.
For a pair $(c,y)$, we approximate $\Delta_{\mathrm{ret}}$ by checking whether $y$ is returned for each question in $\mathcal{Q}(c)$. 
For efficiency and stability, at the start of each epoch, we run the \textsc{MemLife} reader on every training question and cache its memory-access actions. For each candidate $y$, we replay these fixed actions to obtain the returned context $\mathcal{C}_q(y)$ without rerunning reader reasoning. The retrievability reward is
\begin{equation}
    R_{\mathrm{ret}}(c,y)
    =\frac{1}{|\mathcal{Q}(c)|}
    \sum_{q\in\mathcal{Q}(c)}
    \mathbf{1}[y\in \mathcal{C}_q].
    \label{eq:reward_retrievability}
\end{equation}
Since \textsc{FetchMemory} selects entries by timestamp, a segment within a cached fetch interval is returned regardless of its candidate memory, and the other two rewards distinguish such candidates.

\textbf{Multi-granular group-relative optimization.}
For each training segment $c$, the writer samples a group $g=\{y_i\}_{i=1}^{G}$ of $G$ candidates. 
\textsc{MemOpt} combines the three dimensions in FIRM multiplicatively and obtains the reward of token $t$ in candidate $i$ as
\begin{equation}
    x_{i,t}=R_{\mathrm{faith},i,t}R_{\mathrm{inf}}(c,y_i)R_{\mathrm{ret}}(c,y_i).
    \label{eq:memopt_conjunction}
\end{equation}
A token receives high credit only when it is faithful and its memory is informative and retrievable.

Standard group-relative optimization normalizes one scalar reward per candidate and broadcasts the resulting advantage to all of its tokens. \textsc{MemOpt} must instead preserve variation from the token-level faithfulness signal. For candidate $y_i$ with $L_i$ generated tokens, we compute
\begin{equation}
    \bar{x}_i=\frac{1}{L_i}\sum_{t=1}^{L_i}x_{i,t},
    \qquad
    \mu_g=\frac{1}{G}\sum_{i=1}^{G}\bar{x}_i,
    \qquad
    \sigma_g^2=\frac{1}{G}\sum_{i=1}^{G}\frac{1}{L_i}
    \sum_{t=1}^{L_i}(x_{i,t}-\mu_g)^2.
    \label{eq:memopt_group_statistics}
\end{equation}
Averaging within each candidate before computing the group statistics prevents longer memories from dominating the normalization. The token-level advantage is
\begin{equation}
    \widetilde{A}_{i,t}
    =(x_{i,t}-\mu_g) / (\sigma_g+\epsilon_{\mathrm n}),
    \label{eq:memopt_token_advantage}
\end{equation}
where $\epsilon_{\mathrm n}$ stabilizes normalization. Training then follows GRPO \citep{shao2024deepseekmath}.
\section{Experiments}

\subsection{Experimental Setup}

\textbf{Datasets.}
We evaluate on SuperMemory-VQA \citep{alam2026supermemory}, EgoLifeQA \citep{yang2025egolife}, and two extended settings. SuperMemory-LVQA combines all ten SuperMemory-VQA histories while retaining the original test questions, expanding the retrieval space with cross-subject distractors. EgoLife-EQA tests questions about recurring events whose supporting evidence is manually verified against the source recordings. We train \textsc{MemOpt} on SuperMemory-VQA subjects S1--S6, validate on S7--S8, and test on S9--S10. All other benchmarks are used for test only. More details about data are provided in Appendix \ref{app:experimental_details}.

\textbf{Models and baselines.}
Qwen3.5-9B is used as the backbone for both writer and reader across systems. The generalizability study also evaluates Qwen3.6-27B. Training-free baselines include Video ReCap \citep{islam2024videorecap}, EgoRAG \citep{yang2025egolife}, Video-RAG \citep{luo2025videorag}, VideoARM \citep{yin2026videoarm}, EGAgent \citep{rege2026agentic}, and WorldMM \citep{yeo2026worldmm}, which follow their official recommended configurations with all replaceable model components set to Qwen3.5-9B, so that comparisons reflect memory design rather than backbone capacity. Trained baselines include EgoButler \citep{yang2025egolife}, VST \citep{guan2026videostreamingthinking}, TaskMem \citep{zou2026task}, and M3-Agent \citep{long2026seeing}. Appendix \ref{app:implementation_details} provides further details.

Sections \ref{sec:performance}, \ref{sec:memory_quality}, \ref{sec:generalizability}, \ref{sec:ablation_studies} address the following research questions (RQs):
\begin{itemize}
    \item \textbf{RQ1.} Does \textsc{MemLife} outperform existing systems on long-term egocentric video memory question answering? Does \textsc{MemOpt} further improve performance?
    \item \textbf{RQ2.} Do \textsc{MemLife} and \textsc{MemOpt} actually improve memory quality?
    \item \textbf{RQ3.} How generalizable is \textsc{MemOpt} and its trained writer?
    \item \textbf{RQ4.} Is each component in \textsc{MemLife} and \textsc{MemOpt} important?
\end{itemize}
Additional experiments and analyses can be found in the Appendix.

\subsection{Performance Comparison to Baselines}
\label{sec:performance}

Among systems without training, \textsc{MemLife} outperforms baselines in accuracy across benchmarks in Table \ref{tab:main_results}. With two backbones from other model families, \textsc{MemLife} also outperforms EgoRAG and EGAgent (Appendix~\ref{app:stability}). Enabling video access through \textsc{MemLife}-V produces only modest changes, showing that the gains do not depend on revisiting the original recordings. 
On SuperMemory-LVQA, the methods maintain accuracy close to their SuperMemory-VQA results despite lower annotated recall. 
While having cross-subject distractors, SuperMemory-LVQA may also contain subject interactions that provide useful context outside annotated evidence,
which explains the accuracy-recall inconsistency.

\begin{table}[h!] \small
    \caption{
Comparison with existing video memory systems.
Oracle Context bypasses retrieval by supplying all annotated source-video evidence directly to the reader.
\textsc{MemLife}-V permits source-video access.
Bold and underlined values mark the best and second-best results within each group.
}
    \label{tab:main_results}
    \begin{center}
    \resizebox{\linewidth}{!}{
    \begin{tabular}{lcccccccc}
        \toprule
        \multirow{2.5}{*}{Method}
        & \multicolumn{2}{c}{SuperMemory-VQA}
        & \multicolumn{2}{c}{EgoLifeQA}
        & \multicolumn{2}{c}{SuperMemory-LVQA}
        & \multicolumn{2}{c}{EgoLife-EQA} \\
        \cmidrule(lr){2-3}\cmidrule(lr){4-5}\cmidrule(lr){6-7}\cmidrule(l){8-9}
        & Accuracy & Recall
        & Accuracy & Recall
        & Accuracy & Recall
        & Accuracy & Recall \\
        \midrule
        \rowcolor{gray!15}\multicolumn{9}{l}{\textit{Reference}} \\
        \midrule
        Oracle Context & 67.58 & 100.00 & 66.20 & 100.00 & 67.58 & 100.00 & 61.00 & 100.00 \\
        \midrule
        \rowcolor{gray!15}\multicolumn{9}{l}{\textit{Without Memory-Writer Training}} \\
        Video ReCap & 36.28 & 54.01 & 35.20 & 27.40 & 39.17 & 16.79 & 38.00 & 27.00  \\
        EgoRAG & 49.28 & 66.79 & 48.20 & 29.60 & 47.51 & \underline{47.90} & 38.00 & 13.00 \\
        Video-RAG & 43.98 & 59.54 & 38.80 & 20.80 & 49.28 & 35.31 & 28.00 & 6.00 \\
        VideoARM & 39.33 & 70.23 & 36.00 & 45.60 & 40.93 & 9.35 & 40.00 & \textbf{58.00} \\
        EGAgent & 43.66 & 48.09 & 34.00 & 21.80 & 41.73 & 26.53 & 26.00 & 16.00 \\
        WorldMM & 42.05 & 74.62 & 42.20 & 43.80 & 45.91 & 24.05 & 27.00 & 35.00 \\
        \textbf{\textsc{MemLife}} & \underline{56.50} & \underline{80.53} & \underline{52.80} & \underline{48.40} & \underline{56.18} & 46.18 & \textbf{52.00} & {44.00} \\
        \textbf{\textsc{MemLife}-V} & \textbf{57.78} & \textbf{84.35} & \textbf{53.20} & \textbf{50.20} & \textbf{57.95} & \textbf{48.47} & \underline{50.00} & \underline{45.00} \\
        \midrule
        \rowcolor{gray!15}\multicolumn{9}{l}{\textit{With Memory-Writer Training}} \\
        EgoButler & 36.92 & 49.05 & 44.00 & 25.00 & 38.68 & 36.64  & 34.00 & 18.00 \\
        VST & 37.56 & 56.87 & 33.60 & 3.40 & 31.94 & 0.00 & 36.00 & 5.00 \\
        TaskMem & 50.88 & 65.08 & 46.20 & 40.20 & 47.83 & 29.96 & 35.00 & 22.00 \\
        M3-Agent & 43.18 & 31.87 & 30.40 & 7.80 & 43.02 & 10.50 & 34.00 & 4.00 \\
        \textbf{\textsc{MemLife} + \textsc{MemOpt}} & \underline{60.35} & \underline{81.49} & \underline{56.60} & \underline{50.60} & \underline{58.91} & \underline{43.13} & \textbf{57.00} & \underline{41.00} \\
        \textbf{\textsc{MemLife}-V + \textsc{MemOpt}} & \textbf{60.83} & \textbf{83.78} & \textbf{57.60} & \textbf{53.00} & \textbf{61.96} & \textbf{49.62} & \textbf{57.00} & \textbf{43.00} \\
        \bottomrule
    \end{tabular}
    }
    \end{center}
\end{table}

With writer training, \textsc{MemOpt} improves \textsc{MemLife} accuracy on all benchmarks and outperforms every trained baseline. Although trained only on SuperMemory-VQA, it also improves \textsc{MemLife} performance on the other three benchmarks, demonstrating transfer across video distributions, question types, and memory scales. Recall changes are mixed, indicating the accuracy gains also reflect more answer-useful memory content rather than only retrieving annotated evidence.

\subsection{Analysis of Memory Quality}
\label{sec:memory_quality}

\begin{figure}[h!]
    \centering
    \includegraphics[width=1.0\linewidth]{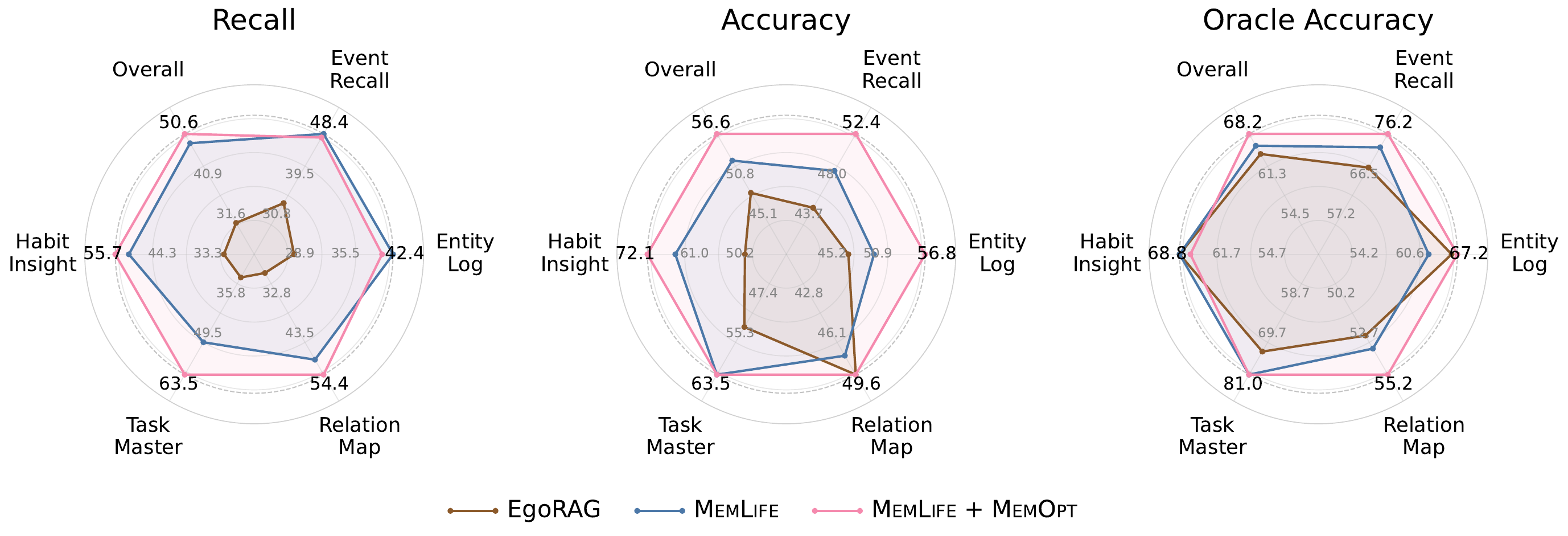}
    \caption{Memory quality across EgoLifeQA question categories. Oracle accuracy is measured by providing memory entries aligned with annotated evidence directly to the reader.}
    \label{fig:memory_quality_analysis_egolife}
\end{figure}

To separate retrieval from the quality of stored memory, Figure \ref{fig:memory_quality_analysis_egolife} reports retrieval recall, standard answer accuracy under normal retrieval, and oracle accuracy when evidence-aligned entries are given directly to the reader. Compared with EgoRAG, \textsc{MemLife} improves recall in every category and raises oracle accuracy overall and in most categories, indicating gains in both retrieval and memory content. \textsc{MemOpt} provides category-dependent gains, improving retrieval for some question types and the answer usefulness of stored content for others. Interestingly, on RelationMap, EgoRAG matches the optimized \textsc{MemLife} system in standard accuracy despite lower recall and oracle accuracy. 
Its normal retrieval may therefore surface alternative useful context outside the annotations, while its evidence-aligned memories do not reliably preserve the relations needed for answering.

Table~\ref{tab:rev_category_results} shows the same analysis on SuperMemory-VQA \citep{alam2026supermemory}, where \textsc{MemLife} + \textsc{MemOpt} outperforms EgoRAG in every category. \textsc{MemOpt} gains most on Conversational Memory and Intent Recall, mainly through more answer-useful content and better retrieval, respectively. Visual Recall shows the largest decrease and has far fewer training question--segment pairs than these two categories (Appendix~\ref{app:training_configuration}). Optimized memories also let the reader answer in fewer rounds (Appendix~\ref{app:efficiency}).

\begin{table}[h!]\small
    \caption{Memory quality across SuperMemory-VQA question categories. Oracle accuracy provides memory entries aligned with annotated evidence directly to the reader.}
    \label{tab:rev_category_results}
    \begin{center}
    \resizebox{\linewidth}{!}{
    \begin{tabular}{llcccccc}
        \toprule
        Metric & Method
        & \makecell{Conversational\\Memory} & \makecell{Intent\\Recall} & \makecell{In-Context\\Retrieval}
        & \makecell{Timeline\\Reconstruction} & \makecell{Object-Location\\Memory} & \makecell{Visual\\Recall} \\
        \midrule
        \multirow{3}{*}{Accuracy}
        & EgoRAG & 69.11 & 70.97 & 31.82 & 45.37 & 40.37 & 34.31 \\
        & \textsc{MemLife} & 63.41 & 77.42 & 43.18 & 57.41 & 52.29 & 44.12 \\
        & \textsc{MemLife} + \textsc{MemOpt} & 79.67 & 88.17 & 45.45 & 56.48 & 50.46 & 39.22 \\
        \midrule
        \multirow{3}{*}{Recall}
        & EgoRAG & 83.67 & 59.68 & 72.41 & 75.51 & 48.28 & 56.52 \\
        & \textsc{MemLife} & 71.43 & 75.81 & 87.36 & 87.76 & 75.86 & 83.70 \\
        & \textsc{MemLife} + \textsc{MemOpt} & 72.45 & 79.03 & 83.91 & 89.80 & 86.21 & 77.17 \\
        \midrule
        \multirow{3}{*}{\makecell[l]{Oracle\\Accuracy}}
        & EgoRAG & 65.85 & 86.02 & 28.41 & 52.78 & 40.37 & 32.35 \\
        & \textsc{MemLife} & 73.98 & 89.25 & 38.64 & 59.26 & 47.71 & 38.24 \\
        & \textsc{MemLife} + \textsc{MemOpt} & 80.49 & 90.32 & 36.36 & 58.33 & 47.71 & 37.25 \\
        \bottomrule
    \end{tabular}
    }
    \end{center}
\end{table}

Table \ref{tab:memory_quality_case} illustrates how \textsc{MemOpt} changes the stored content. In the first example, the untrained writer mistakes lentils for corn, while \textsc{MemOpt} corrects the object without losing the surrounding action. In the second, the original memory uses a vague pronoun and omits the relevant food, whereas \textsc{MemOpt} identifies the person and records the baked chicken needed to answer the question. These examples show that \textsc{MemOpt} removes unsupported details while making answer-relevant entities and events explicit, improving both faithfulness and informativeness.

\begin{table}[h!] \small
    \centering
    \caption{Examples of memory corrections learned through \textsc{MemOpt}. Red highlights errors and yellow highlights corrected content.}
    \begin{center}
    \begin{tabular}{@{}m{0.10\linewidth}m{0.40\linewidth}m{0.40\linewidth}@{}}
    \toprule
        Frames
        & \includegraphics[width=0.32\linewidth]{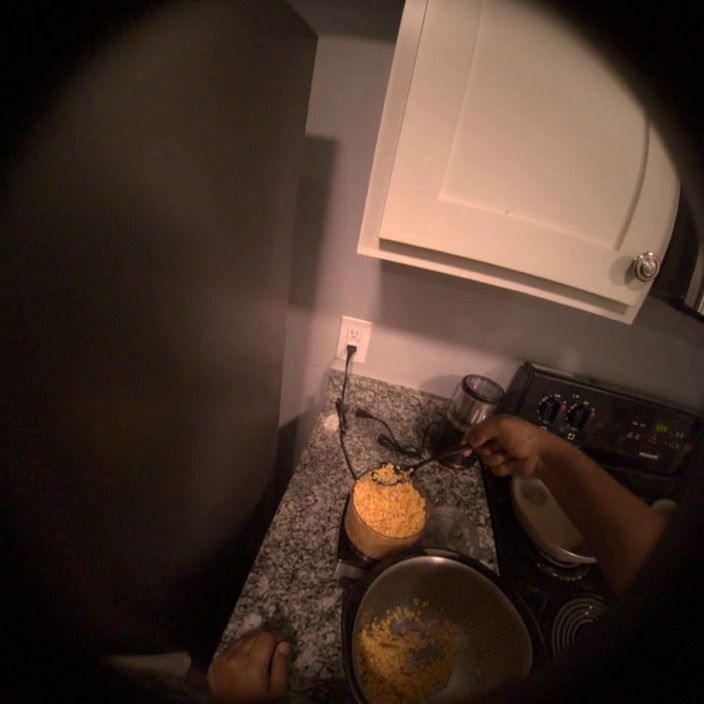}\hfill
          \includegraphics[width=0.32\linewidth]{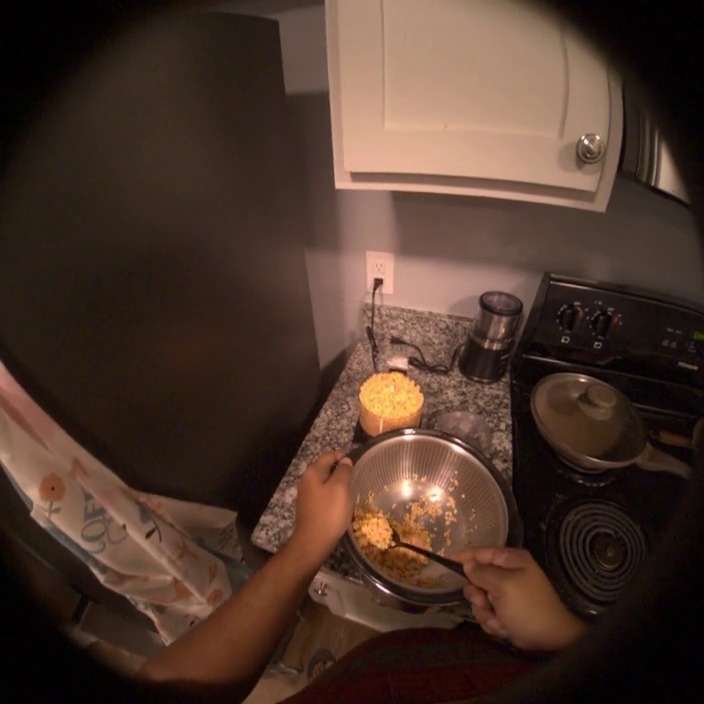}\hfill
          \includegraphics[width=0.32\linewidth]{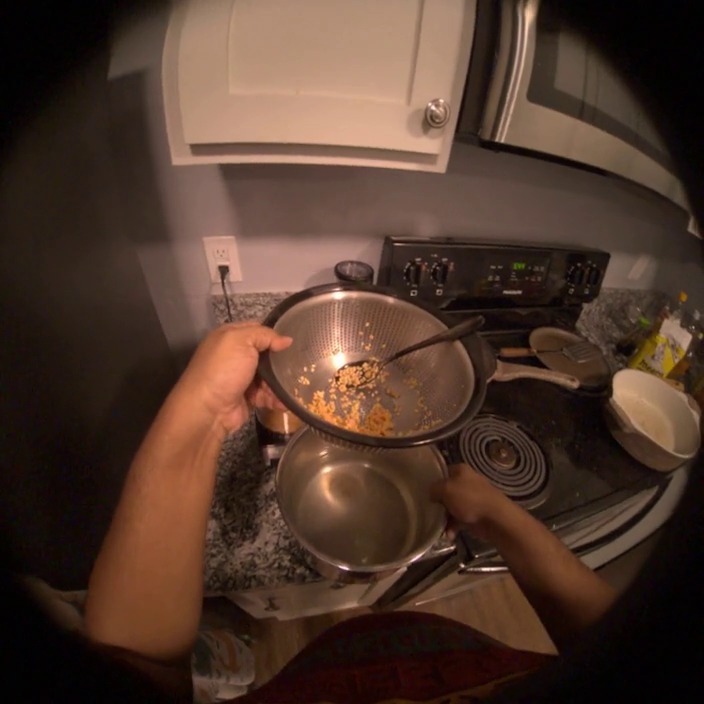}
        & \includegraphics[width=0.32\linewidth]{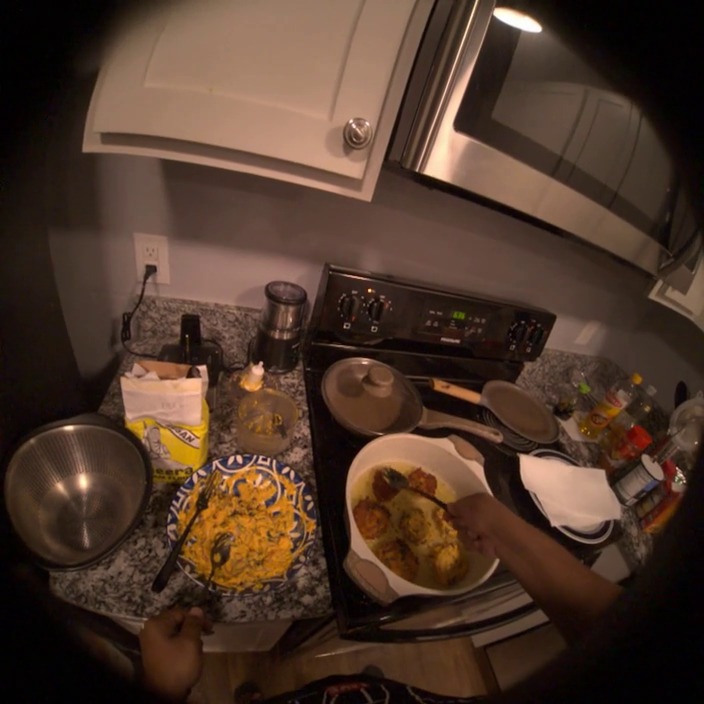}\hfill
          \includegraphics[width=0.32\linewidth]{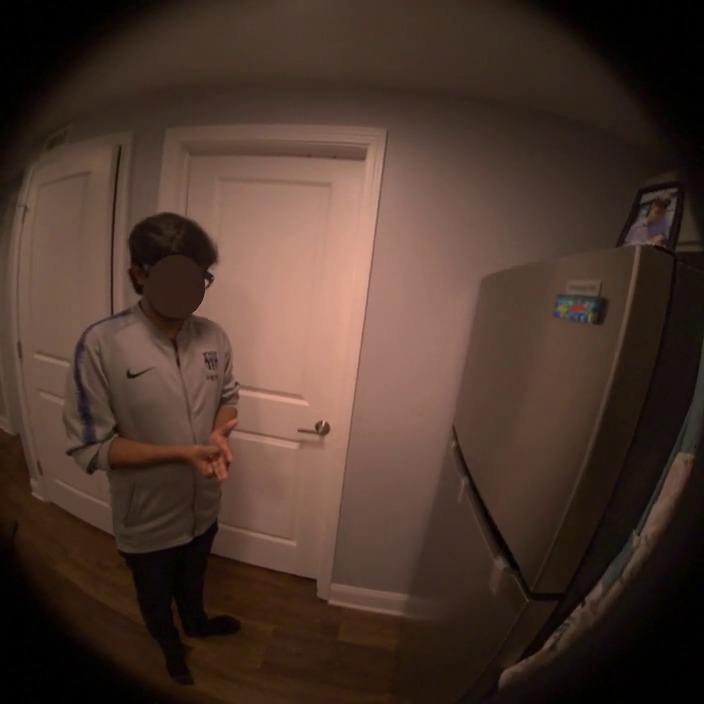}\hfill
          \includegraphics[width=0.32\linewidth]{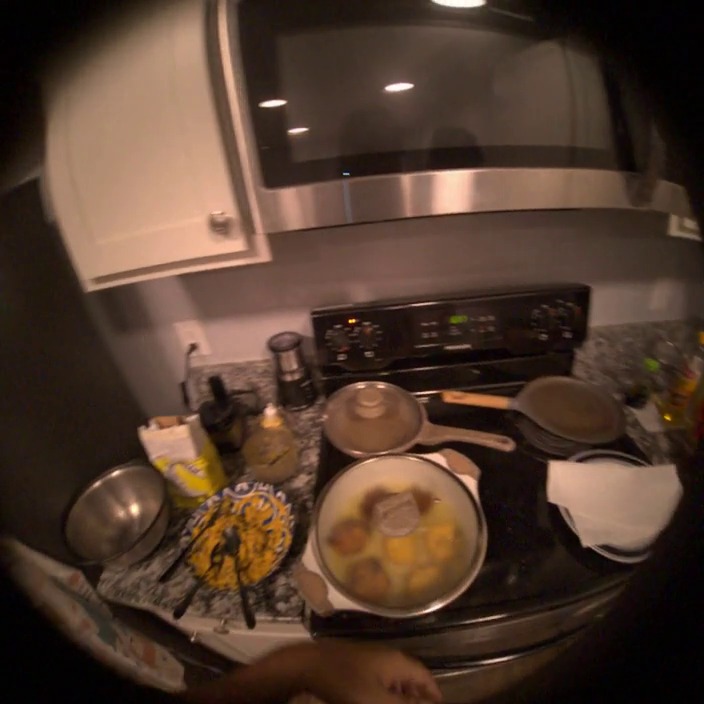} \\
    \midrule
        \multirow{3}{*}{Question}
        & Q: \textit{[...] Did I add the milk before or after the eggs when making the batter?}
        & Q: \textit{[...] Did I set a reminder for a backup dinner plan?} \\
        & A: \textit{You did not add milk or eggs to the batter; you only added lentils, salt, and spices.}
        & A: \textit{No [...] However, you did mention earlier that you have baked chicken available.} \\
    \midrule
        \textsc{MemLife}
        & I am in a kitchen [...] transfer \sethlcolor{lightred}\hl{yellow corn kernels} from a small food processor bowl [...] move toward the sink area [...]
        & [...] preparing food [...] \sethlcolor{lightred}\hl{they} respond to my question about being hungry by saying they can wait for the food [...] \\
    \midrule
        \makecell[l]{\textsc{MemLife}\\+ \textsc{MemOpt}}
        & I am in a kitchen  [...] yellowish-orange granular material, which appears to be \hl{cooked lentils} [...] to the sink area [...]
        & [...] I am preparing a meal, specifically baked chicken, and offering it to \hl{B}, who indicates they are not hungry and can wait. \\
    \bottomrule
    \end{tabular}
    \end{center}
    \label{tab:memory_quality_case}
\end{table}

\subsection{Generalizability of \textsc{MemOpt} Training}
\label{sec:generalizability}

We then examine whether \textsc{MemOpt} depends on the writer backbone by optimizing both Qwen3.5-9B and Qwen3.6-27B. Table \ref{tab:writer_reader_backbones} shows that \textsc{MemOpt} improves every writer--reader pairing on both benchmarks. Changing the writer scale produces only modest differences, indicating that the training benefit does not depend on a particular writer backbone.

The Qwen3.6-27B reader further tests whether the optimized writer transfers beyond the Qwen3.5-9B reader used to collect retrievability supervision. Using the stronger reader substantially raises absolute accuracy for both writers. Under greedy decoding, \textsc{MemOpt} still improves every setting, with smaller gains that depend on how far zero-shot memory leaves the reader below oracle context (Appendix~\ref{app:stability}). The best results combine the stronger reader with an optimized writer, indicating that reader scaling and writer optimization are complementary.

\begin{table}[h!]\small
    \caption{Answer accuracy (\%) across writer and reader backbones.}
    \label{tab:writer_reader_backbones}
    \begin{center}
    \begin{tabular}{llcccc}
        \toprule
        \multirow{2.5}{*}{Writer} & \multirow{2.5}{*}{Training}
        & \multicolumn{2}{c}{Qwen3.5-9B Reader}
        & \multicolumn{2}{c}{Qwen3.6-27B Reader} \\
        \cmidrule(lr){3-4}\cmidrule(l){5-6}
        & & SuperMemory-VQA & EgoLifeQA 
        & SuperMemory-VQA & EgoLifeQA \\
        \midrule
        \multirow{2}{*}{Qwen3.5-9B}
        & Zero-shot & 56.50 & 52.80 & 66.93 & 59.60 \\
        & \textsc{MemOpt} & \textbf{60.35} & \textbf{56.60} & \textbf{67.90} & \textbf{60.20} \\
        \midrule
        \multirow{2}{*}{Qwen3.6-27B}
        & Zero-shot & 57.14 & 55.20 & 66.77 & 59.80 \\
        & \textsc{MemOpt} & \textbf{60.67} & \textbf{56.60} & \textbf{68.06} & \textbf{60.60} \\
        \bottomrule
    \end{tabular}
    \end{center}
\end{table}

\begin{wrapfigure}{r}{0.63\linewidth}
   \centering
   \includegraphics[width=\linewidth]{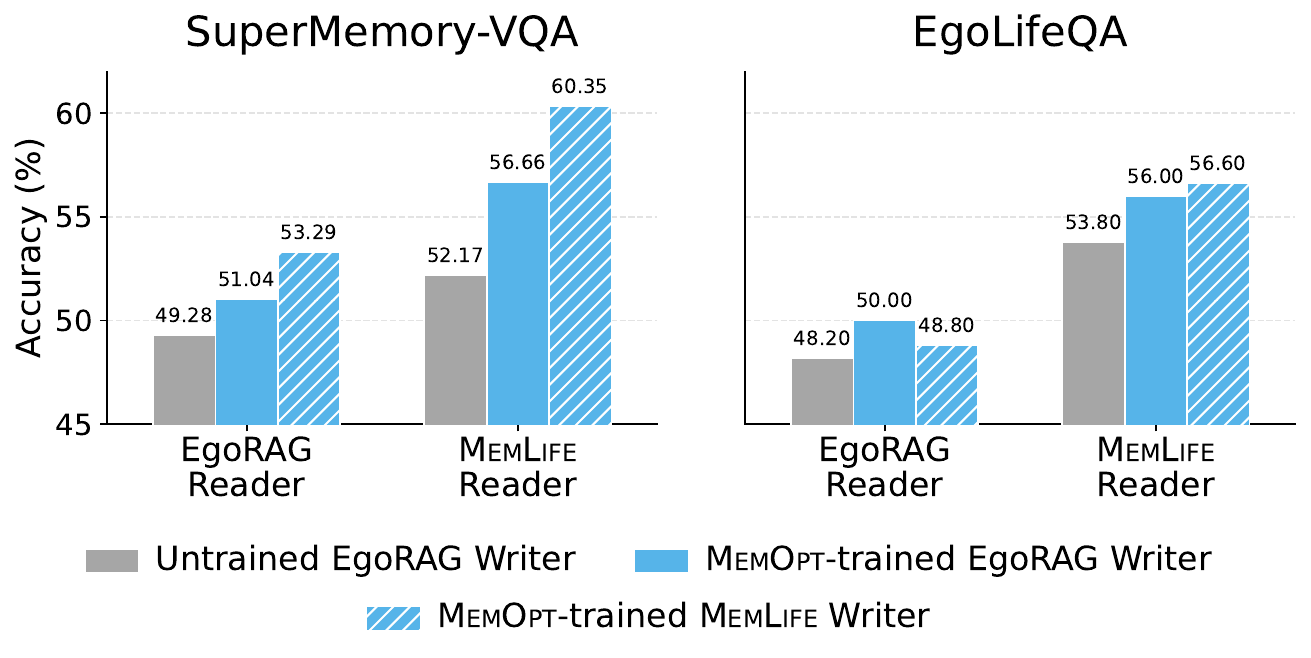}
   \caption{Generalizability of \textsc{MemOpt} writer training across memory systems.}
   \vspace{-0.1in}
   \label{fig:writer_generalizability}
\end{wrapfigure}

Beyond backbone changes, we study whether \textsc{MemOpt} transfers across memory systems. We optimize the EgoRAG writer and evaluate the original and optimized memories with both the native EgoRAG reader and the \textsc{MemLife} agentic reader. We also evaluate the optimized \textsc{MemLife} writer with the EgoRAG reader. 
Figure \ref{fig:writer_generalizability} shows that training improves the native EgoRAG pipeline on both benchmarks.
Using \textsc{MemLife} to read optimized EgoRAG memories provides further gains, while 
the highest performance is achieved with the trained \textsc{MemLife} writer.

Since retrievability is optimized with first-person training questions, we also test robustness to question phrasing. We use Qwen3.6-27B to rewrite the test questions into a third-person form and a neutral form without personal references, and evaluate the same memories with the 9B reader. Table~\ref{tab:rev_question_phrasing} shows that \textsc{MemOpt} improves \textsc{MemLife} under both rewrites on both benchmarks, and the advantage of \textsc{MemLife} over EgoRAG does not shrink. The gains of \textsc{MemOpt} are largest on the original first-person questions, which match its training questions as well as the way users naturally ask about their own experiences.

\begin{table}[h!]\small
    \caption{Answer accuracy (\%) under rewritten question phrasings. All methods use the same memories and the Qwen3.5-9B reader.}
    \label{tab:rev_question_phrasing}
    \begin{center}
    \resizebox{\linewidth}{!}{
    \begin{tabular}{lcccccc}
        \toprule
        \multirow{2.5}{*}{Method}
        & \multicolumn{3}{c}{SuperMemory-VQA}
        & \multicolumn{3}{c}{EgoLifeQA} \\
        \cmidrule(lr){2-4}\cmidrule(l){5-7}
        & First-person & Third-person & Neutral & First-person & Third-person & Neutral \\
        \midrule
        EgoRAG & 49.28 & 48.48 & 48.15 & 48.20 & 45.20 & 44.80 \\
        \textsc{MemLife} & 56.50 & 56.98 & 57.95 & 52.80 & 55.40 & 52.80 \\
        \textsc{MemLife} + \textsc{MemOpt} & \textbf{60.35} & \textbf{57.30} & \textbf{59.55} & \textbf{56.60} & \textbf{56.80} & \textbf{54.80} \\
        \bottomrule
    \end{tabular}
    }
    \end{center}
\end{table}

\subsection{Ablation Studies} \label{sec:ablation_studies}

To analyze how different components in our proposed methods contribute to the overall performance, we first ablate each component in the writer and reader designs of \textsc{MemLife}, with the results shown in Table \ref{tab:writer_reader_ablation}.
For the writer design, we progressively add entity grounding and first-person narration to multimodal fusion, and examine how performance will change.

\begin{table}[h!]\small
    \caption{Ablation studies on the writer and reader components in \textsc{MemLife}. Accuracy is reported for SuperMemory-VQA and EgoLifeQA. Recall on the long-term EgoLifeQA task is also reported.}
    \label{tab:writer_reader_ablation}
    \begin{center}
    \resizebox{\linewidth}{!}{
    \begin{tabular}{ccccccccc}
        \toprule
        \rowcolor{gray!15}\multicolumn{9}{c}{\textit{Writer Ablation}} \\
        \midrule
        \multirow{2.5}{*}{\makecell{Multimodal\\Fusion}}
        & \multirow{2.5}{*}{\makecell{Entity\\Grounding}}
        & \multirow{2.5}{*}{\makecell{First-person\\Narration}}
        & \multicolumn{2}{c}{SuperMemory-VQA}
        & \multicolumn{2}{c}{EgoLifeQA}
        & \multicolumn{2}{c}{EgoLifeQA Recall} \\
        \cmidrule(lr){4-5}\cmidrule(l){6-7}\cmidrule(l){8-9}
        & & & Zero-shot & \textsc{MemOpt}
        & Zero-shot & \textsc{MemOpt}
        & Zero-shot & \textsc{MemOpt} \\
        \midrule
        \greencheck & \redcross   & \redcross   & 52.33 & 56.98  & 52.80 & 52.60 & 48.60 & 47.00 \\
        \greencheck & \greencheck & \redcross   & 54.09 & 59.87 & 52.00 & 54.40 & 45.60 & 47.60 \\
        \greencheck & \greencheck & \greencheck & 56.50 & 60.35 & 52.80 & 56.60 & 48.40 & 50.60 \\
        \midrule
        \rowcolor{gray!15}\multicolumn{9}{c}{\textit{Reader Ablation}} \\
        \midrule
        \multirow{2.5}{*}{\makecell{Agentic\\Reasoning}}
        & \multirow{2.5}{*}{\makecell{Time\\Anchoring}}
        & \multirow{2.5}{*}{\makecell{Chronological\\Ordering}}
        & \multicolumn{2}{c}{SuperMemory-VQA}
        & \multicolumn{2}{c}{EgoLifeQA}
        & \multicolumn{2}{c}{EgoLifeQA Recall} \\
        \cmidrule(lr){4-5}\cmidrule(l){6-7}\cmidrule(l){8-9}
        & & & Zero-shot & \textsc{MemOpt}
        & Zero-shot & \textsc{MemOpt}
        & Zero-shot & \textsc{MemOpt} \\
        \midrule
        \greencheck & \redcross   & \redcross   & 56.02 & 58.27  & 50.40 & 51.80 & 50.60 & 48.20 \\
        \greencheck & \greencheck & \redcross   & 56.50 & 58.59 & 50.60 & 54.00 & 50.00 & 50.00 \\
        \greencheck & \greencheck & \greencheck & 56.50 & 60.35 & 52.80 & 56.60 & 48.40 & 50.60 \\
        \bottomrule
    \end{tabular}
    }
    \end{center}
\end{table}

From the upper block of Table \ref{tab:writer_reader_ablation}, we observe that entity grounding generally improves accuracy, particularly after \textsc{MemOpt} training, but provides little benefit to recall.
First-person narration further improves accuracy and consistently raises EgoLifeQA recall, supporting its role in aligning written memories with wearer-centered queries.

We then ablate the \textsc{MemLife} reader by adding time anchoring and chronological ordering to agentic reasoning. The lower block of Table \ref{tab:writer_reader_ablation} shows that 
having time anchoring in the search tool provides its clearest benefit on optimized EgoLifeQA memories, where its accuracy gain is accompanied by higher recall.
Chronological ordering further improves accuracy despite small or mixed recall changes, indicating that preserving event order primarily benefits reasoning over retrieved evidence. The complete reader performs best overall, with the largest gains appearing after writer optimization.

For the design of \textsc{MemOpt}, we compare various supervision signals used to train the writer. Table \ref{tab:reward_ablation} shows that teacher imitation with Qwen3.6-27B provides only modest gains, while final-answer accuracy supervision produces inconsistent changes across benchmarks. Among the proposed reward dimensions, retrievability alone raises recall on both benchmarks but does not consistently improve accuracy. Adding informativeness strongly benefits SuperMemory-VQA, although its gain does not transfer to EgoLifeQA. With faithfulness as a regularizer, the complete objective achieves the highest accuracy on both benchmarks while retaining higher recall than the untrained writer.

\begin{table}[h!]\small
    \caption{Writer supervision signals. Accuracy rewards each segment with the answer accuracy of its questions, using the same annotations as FIRM. Teacher imitates Qwen3.6-27B memories.}
    \label{tab:reward_ablation}
    \begin{center}
    \resizebox{\linewidth}{!}{
    \begin{tabular}{cccccccccc}
        \toprule
        \multicolumn{5}{c}{Writer Supervision}
        & \multicolumn{2}{c}{SuperMemory-VQA}
        & \multicolumn{2}{c}{EgoLifeQA} \\
        \cmidrule(lr){1-5}\cmidrule(lr){6-7}\cmidrule(l){8-9}
        Teacher & Accuracy & Retrievability & Informativeness & Faithfulness
        & Accuracy & Recall & Accuracy & Recall \\
        \midrule
        \redcross & \redcross & \redcross & \redcross & \redcross & 56.50 & 80.53 & 52.80 & 48.40 \\
        \greencheck & \redcross & \redcross & \redcross & \redcross & 57.78 & 83.59 & 53.00 & 49.00 \\
        \redcross & \greencheck & \redcross & \redcross & \redcross & 57.46 & 79.01 & 52.60 & 47.80 \\
        \redcross & \redcross & \greencheck & \redcross & \redcross & 53.45 & 82.63 & 54.40 & \textbf{53.20} \\
        \redcross & \redcross & \greencheck & \greencheck & \redcross & 60.03 & \textbf{84.35} & 51.40 & 47.60 \\
        \redcross & \redcross & \greencheck & \greencheck & \greencheck & \textbf{60.35} & 81.49 & \textbf{56.60} & 50.60 \\
        \bottomrule
    \end{tabular}
    }
    \end{center}
\end{table}

Finally, we perform ablation studies on the faithfulness granularity and reward aggregation strategy used in \textsc{MemOpt}. Following the probability-based judgment used for informativeness, the sequence-level variant assigns every token the same score about whether the complete memory is supported by its source segment. Different from multiplicative aggregation, the additive variant sums the three rewards. As shown in Table \ref{tab:additional_reward_ablation}, under additive aggregation, token-level faithfulness maintains similar SuperMemory-VQA accuracy with slightly lower recall, but improves both metrics on out-of-domain EgoLifeQA. With token-level faithfulness fixed, multiplicative aggregation further improves accuracy on both benchmarks, and achieves the best performance overall. 

\begin{table}[h!]\small
    \caption{Ablation of faithfulness granularity and reward aggregation.}
    \label{tab:additional_reward_ablation}
    \begin{center}
    \begin{tabular}{llcccc}
        \toprule
        \multicolumn{2}{c}{Training Design}
        & \multicolumn{2}{c}{SuperMemory-VQA}
        & \multicolumn{2}{c}{EgoLifeQA} \\
        \cmidrule(lr){1-2}\cmidrule(lr){3-4}\cmidrule(l){5-6}
        Faithfulness & Aggregation & Accuracy & Recall & Accuracy & Recall \\
        \midrule
        Sequence-level  & Additive       & 58.75 & 83.78 & 55.00 & 46.40 \\
        Token-level & Additive       & 58.91 & 81.68 & 56.00 & 49.40 \\
        Token-level & Multiplicative & 60.35 & 81.49 & 56.60 & 50.60 \\
        \bottomrule
    \end{tabular}
    \end{center}
\end{table}
\section{Conclusion}

We introduced \textsc{MemLife}, an agentic memory system for long-term egocentric video that constructs time- and entity-anchored, first-person episodes and accesses them through time-scoped retrieval and chronological evidence organization.
We further proposed \textsc{MemOpt}, which applies reinforcement learning only to the memory writer through the FIRM objective for faithful, informative, and retrievable memories.
\textsc{MemLife} outperforms state-of-the-art training-free systems, while \textsc{MemOpt} provides further gains that transfer across writer and reader backbones, memory systems, and out-of-domain video and question distributions.
Together, the results demonstrate that learning what to remember is an effective and generalizable approach to long-term video question answering.

\clearpage
\newpage
\bibliographystyle{assets/plainnat}
\bibliography{paper}

\clearpage
\newpage
\beginappendix

\section{Dataset and Baseline Details}
\label{app:experimental_details}

\subsection{Datasets and Splits}
\label{app:datasets}

SuperMemory-VQA contains ten subjects recorded across multiple sessions \citep{alam2026supermemory}. We exclude 82 questions whose annotated evidence refers to source videos unavailable in the public release, leaving 4,771 questions. We partition these questions by subject, using S1--S6 for training, S7--S8 for validation, and S9--S10 for testing. These splits contain 3,425, 723, and 623 questions, respectively. EgoLifeQA contains 500 multiple-choice questions about a continuous seven-day recording \citep{yang2025egolife}. We use its complete question set only for testing.

SuperMemory-LVQA increases the retrieval space by concatenating the histories of all ten SuperMemory-VQA subjects into one memory store. It retains the 623 questions from the SuperMemory-VQA test split, with the remaining subject histories as additional distractors. 

EgoLife-EQA contains 100 test questions about recurring events across multiple days in EgoLife. We identify candidate events from the official captions and formulate questions about them. Two annotators independently verify whether each candidate evidence segment supports its question's annotated answer, reaching 92.1\% agreement. They resolve disagreements through discussion, and we discard questions without verified evidence or an unambiguous answer.

Table~\ref{tab:egolife_eqa_types} groups the EgoLife-EQA questions into three categories. Frequency-counting questions ask on which days an activity occurred and may have multiple correct options, such as ``\emph{On which days during the seven-day period did I go grocery shopping? Choose all that apply.}'' Routine questions ask what activity typically occurs at a given time of day, such as ``\emph{What do I usually check on my phone in the morning?}'' Comparison questions ask which of two activities occurred on more days, such as ``\emph{Which activity did I do in the bedroom on more days---browsing social media or browsing products online?}''

\begin{table}[h!] \small
    \caption{Composition of EgoLife-EQA by question type. Correct options, evidence segments, and distinct evidence days are averaged per question.}
    \label{tab:egolife_eqa_types}
    \begin{center}
\begin{tabular}{lccccc}
    \toprule
    Type & \makecell{\# Questions} & \makecell{\# Options} & \makecell{Avg. Correct\\Options} & \makecell{Avg. Evidence\\Segments} & \makecell{Avg. Distinct\\Days} \\
    \midrule
    Frequency counting & 45 & 8 & 2.33 & 2.31 & 2.31 \\
    Routine            & 31 & 2--4 & 1.00 & 2.97 & 2.97 \\
    Comparison         & 24 & 3 & 1.00 & 8.04 & 5.54 \\
    \midrule
    All                & 100 & -- & 1.60 & 3.89 & 3.29 \\
    \bottomrule
\end{tabular}
    \end{center}
\end{table}

Table~\ref{tab:dataset_statistics} compares the scale of the accessible video history and evidence annotations across the four evaluation settings, where only SuperMemory-VQA supplies supervision for \textsc{MemOpt}.

\begin{table}[h!] \small
    \caption{Dataset statistics. Video hours and time spans are averaged over the history available to each question. Evidence segments and their total duration are averaged per question.}
    \label{tab:dataset_statistics}
    \begin{center}
\begin{tabular}{lccccc}
    \toprule
    Dataset & \makecell{\# Questions} & \makecell{Avg. Video\\Hours} & \makecell{Avg. Time\\Span (days)}
    & \makecell{Avg. Evidence\\Segments} & \makecell{Avg. Evidence\\Duration (s)} \\
    \midrule
    SuperMemory-VQA  & 4771 & 3.97  & 12.50  & 1.34 & 57.1  \\
    EgoLifeQA        & 500  & 22.65 & 2.80   & 1.10 & 32.3  \\
    SuperMemory-LVQA & 623  & 47.90 & 127.36 & 1.24 & 56.3  \\
    EgoLife-EQA      & 100  & 43.06 & 6.33   & 3.89 & 116.7 \\
    \bottomrule
\end{tabular}
    \end{center}
\end{table}

\subsection{Baselines}
\label{app:baselines}

\paragraph{Without task-specific writer training.}
Video ReCap recursively builds clip-, segment-, and video-level captions by combining visual features with captions from the preceding hierarchy \citep{islam2024videorecap}. EgoRAG builds clip-, hour-, and day-level memories and retrieves relevant clips using visual and textual similarity \citep{yang2025egolife}. Video-RAG indexes OCR, ASR, and object-detection text, then provides retrieved text and sampled video frames to a VLM \citep{luo2025videorag}. The remaining systems perform adaptive multimodal access. VideoARM constructs a query-conditioned hierarchical memory online while inspecting progressively narrower regions of the source video \citep{yin2026videoarm}. EGAgent plans over a temporal entity scene graph with visual-frame and transcript search \citep{rege2026agentic}. WorldMM iteratively retrieves from multiscale episodic and semantic graphs and a visual memory \citep{yeo2026worldmm}.

\paragraph{With learned memory construction.}
EgoButler combines EgoRAG with EgoGPT, which is fine-tuned for both visual-audio captioning and question answering \citep{yang2025egolife}. VST trains a single streaming VideoLLM to generate both textual memory and final answers through supervised fine-tuning and answer-based reinforcement learning \citep{guan2026videostreamingthinking}. TaskMem instead optimizes a memorization policy with model-judged quality rewards followed by task-relevance preference learning, leaving QA to a separate answer generator \citep{zou2026task}. M3-Agent trains an entity-centric episodic and semantic memory writer through imitation of synthetic demonstrations, then separately trains its memory-search controller with reinforcement learning \citep{long2026seeing}.

\paragraph{Evaluation protocol.}
For EgoButler, VST, TaskMem, and M3-Agent, we use the released EgoGPT-7B, VST-32B, TaskMem-30B, and M3-Agent memorization and control checkpoints, respectively,\footnote{Official checkpoints: \href{https://huggingface.co/lmms-lab/EgoGPT-7b-EgoIT-EgoLife}{EgoGPT-7B}, \href{https://huggingface.co/Catalan258/VST-32B}{VST-32B}, \href{https://huggingface.co/ByteDance-Seed/TaskMem}{TaskMem-30B}, \href{https://huggingface.co/ByteDance-Seed/M3-Agent-Memorization}{M3-Agent-Memorization}, and \href{https://huggingface.co/ByteDance-Seed/M3-Agent-Control}{M3-Agent-Control}.} where some have much larger capacity than our tested 9B model.
All other replaceable model components use Qwen3.5-9B. Appendix \ref{app:controlled_trained_baselines} shows additional experimental results on the reproduced training methods with matched backbone models and training data. Apart from this backbone change, the training-free baselines follow their official recommended configurations, so that the comparison reflects differences in memory design rather than in backbone capacity. Our evaluation of the released TaskMem checkpoint reaches EgoLifeQA accuracy close to that reported by \citet{zou2026task}, supporting the validity of our evaluation setup. Compared with the released checkpoints, our Qwen3.5-9B reproductions in Appendix~\ref{app:controlled_trained_baselines} perform better on some benchmarks and worse on others, indicating that the reproduction does not systematically disadvantage these baselines.

All methods receive the same questions, answer options, and causal video histories, and we recompute their metrics using a common answer parser. The \emph{Oracle Context} reference in Table \ref{tab:main_results} bypasses retrieval by directly supplying the answerer with all annotated source-video evidence for each question. Its retrieval recall is therefore 100\% by construction. Among baselines without task-specific writer training, Video-RAG, VideoARM, EGAgent, and WorldMM retain query-time access to visual evidence, whereas the default \textsc{MemLife} operates only on written memory. Among learned systems, we implement VST and M3-Agent with models trained on both memory construction and their answering or control components. EgoButler, TaskMem, and \textsc{MemOpt} instead adopt only a trained memory writer while keeping the downstream reader fixed.

\paragraph{Design comparison.} Table~\ref{tab:rev_design_comparison} compares key designs adopted by \textsc{MemLife} and the evaluated training-free systems. Each baseline lacks at least two of the six designs, whereas \textsc{MemLife} combines all of them. Table~\ref{tab:egagent_comparison} further compares \textsc{MemLife} with EGAgent, which shares all of these designs except first-person narration and chronological ordering. Without these two designs, \textsc{MemLife} still clearly outperforms EGAgent. Unlike \textsc{MemLife}, which answers questions from its episodic memory, EGAgent uses its captions only to build an entity graph, where it performs entity grounding. Its reader relies on this graph and the raw videos without access to the captions. Running \textsc{MemLife}'s reader on these captions substantially improves accuracy, even though they lack the grounding performed in the graph. Replacing them with \textsc{MemLife}'s own entries, which ground entities during writing, raises accuracy further, and chronological ordering, first-person narration, and \textsc{MemOpt} each bring additional gains.

\textsc{MemLife} provides the empirical contribution of our work by showing what practical designs such as those compared above can achieve when combined in one system. Building on this system, \textsc{MemOpt} provides the methodological contribution by optimizing what is written into memory, with rewards that directly measure the quality of each entry. Both contributions point to the written memory itself as a strong lever for egocentric video QA. By reading a well-designed memory directly, \textsc{MemLife} already achieves higher accuracy than every baseline in Table~\ref{tab:main_results} without any training, including systems that build graphs, revisit raw videos, or train their components. With the reader fixed, \textsc{MemOpt} raises accuracy on all four benchmarks and also improves other memory systems (Figure~\ref{fig:writer_generalizability}).

\begin{table}[h!]\small
    \caption{Key designs combined in \textsc{MemLife} and in the evaluated training-free systems.}
    \label{tab:rev_design_comparison}
    \begin{center}
    \begin{tabular}{lcccccc}
        \toprule
        Method & \makecell{Multimodal\\Fusion} & \makecell{Entity\\Grounding} & \makecell{First-Person\\Narration}
        & \makecell{Agentic\\Reasoning} & \makecell{Time\\Anchoring} & \makecell{Chronological\\Ordering} \\
        \midrule
        Video ReCap & \redcross & \redcross & \redcross & \redcross & \redcross & \redcross \\
        EgoRAG & \greencheck & \redcross & \greencheck & \redcross & \greencheck & \redcross \\
        Video-RAG & \redcross & \redcross & \redcross & \redcross & \redcross & \greencheck \\
        VideoARM & \redcross & \redcross & \redcross & \greencheck & \greencheck & \redcross \\
        EGAgent & \greencheck & \greencheck & \redcross & \greencheck & \greencheck & \redcross \\
        WorldMM & \greencheck & \redcross & \redcross & \greencheck & \greencheck & \redcross \\
        \textsc{MemLife} & \greencheck & \greencheck & \greencheck & \greencheck & \greencheck & \greencheck \\
        \bottomrule
    \end{tabular}
    \end{center}
\end{table}

\begin{table}[h!]\small
    \caption{Accuracy of EGAgent and \textsc{MemLife} variants, all with the same Qwen3.5-9B backbone. Memory shows whose captions or entries are used. Rows with episodic memory access use \textsc{MemLife}'s reader.}
    \label{tab:egagent_comparison}
    \begin{center}
    {\setlength{\tabcolsep}{3pt}
    \begin{tabular}{lccccccc}
        \toprule
        Memory & \makecell{Episodic\\Memory\\Access} & \makecell{Entity\\Grounding} & \makecell{Chronological\\Ordering} & \makecell{First-Person\\Narration} & \textsc{MemOpt} & \makecell{SuperMemory-\\VQA} & EgoLifeQA \\
        \midrule
        EGAgent & \redcross & \greencheck & \redcross & \redcross & \redcross & 43.66 & 34.00 \\
        EGAgent & \greencheck & \redcross & \redcross & \redcross & \redcross & 51.36 & 50.60 \\
        \textsc{MemLife} & \greencheck & \greencheck & \redcross & \redcross & \redcross & 52.17 & 50.80 \\
        \textsc{MemLife} & \greencheck & \greencheck & \greencheck & \redcross & \redcross & 54.09 & 52.00 \\
        \textsc{MemLife} & \greencheck & \greencheck & \greencheck & \greencheck & \redcross & 56.50 & 52.80 \\
        \textsc{MemLife} & \greencheck & \greencheck & \greencheck & \greencheck & \greencheck & 60.35 & 56.60 \\
        \bottomrule
    \end{tabular}}
    \end{center}
\end{table}

\subsection{Evaluation Metrics}
\label{app:evaluation_metrics}

Let $\mathcal{Q}$ denote the test questions, $A_q$ the set of annotated correct option labels, and $\widehat{A}_q$ the predicted set produced by the answer parser. These sets contain one label for single-choice questions. For multi-select EgoLife-EQA questions, a prediction is correct only if it exactly matches the complete annotated set. We compute answer accuracy over all test questions as
\begin{equation}
    \operatorname{Acc}
    =\frac{1}{|\mathcal{Q}|}
    \sum_{q\in\mathcal{Q}}\mathbf{1}[\widehat{A}_q=A_q].
    \label{eq:evaluation_accuracy}
\end{equation}

Retrieval recall is computed over the subset $\mathcal{Q}_G\subseteq\mathcal{Q}$ containing questions with at least one annotated temporal evidence interval. Let $G_q$ and $H_q$ denote the source-indexed temporal intervals annotated for $q$ and returned to the reader, respectively. A question counts as retrieved when at least one returned interval overlaps an annotated interval from the same source video. We compute
\begin{equation}
    \operatorname{Recall}=\frac{1}{|\mathcal{Q}_{G}|}
    \sum_{q\in\mathcal{Q}_{G}}
    \mathbf{1}\!\left[\exists g\in G_q,\,h\in H_q\ \text{such that}\ h\cap g\neq\varnothing\right].
    \label{eq:evaluation_recall}
\end{equation}
For a single-shot reader, $H_q$ is its retrieved context. For an agentic reader, $H_q$ is the union of evidence returned by all executed search and fetch actions. This union measures the evidence actually available during the interaction rather than evidence recoverable by an unexecuted query.

\section{Implementation Details}
\label{app:implementation_details}

\subsection{Models and Inference}
\label{app:inference_details}

Both the default \textsc{MemLife} writer and reader use Qwen3.5-9B. The backbone study additionally uses Qwen3.6-27B as the writer and reader, producing all four combinations of the two models. Each writer generates one memory store per dataset, and that same store is evaluated by both reader backbones. Swapping the reader therefore does not regenerate or alter the memory. All reader weights remain frozen, including during \textsc{MemOpt} training.

The writer processes each 30-second segment independently from eight frames at 704-pixel resolution and its speech transcript. A trailing segment shorter than one second is folded into the preceding segment. Memory descriptions are generated greedily with a maximum of 512 tokens. We embed each description using BAAI/bge-large-en-v1.5 and perform exact inner-product search over normalized embeddings.

At evaluation, the agentic reader can call \textsc{SearchMemory} and \textsc{FetchMemory} for at most ten rounds. Semantic search returns at most 32 entries, while interval-based fetching returns at most 64 entries. The default reader has no access to source video. The \textsc{MemLife}-V variant additionally allows the reader to inspect up to 50 sampled frames and the transcript from a selected temporal interval. Both writer and reader generations use greedy decoding for deterministic results. More details about the writer and reader prompts are in Appendix \ref{app:prompt_and_tool_details}.

\subsection{\textsc{MemOpt} Training}
\label{app:training_configuration}

\paragraph{Training configuration.} All evaluators use Qwen3.5-9B, and both their parameters and the reader parameters remain frozen. We train the writer for three epochs and rebuild the validation memory after each epoch. We select the checkpoint with the highest sum of answer accuracy and retrieval recall on S7--S8, then evaluate it once on each test benchmark. Table~\ref{tab:memopt_hyperparameters} summarizes the training configuration.

The Kullback--Leibler regularizer is applied directly to the loss against the frozen initial writer rather than incorporated into the reward, so it does not affect the group-relative advantages. Training rollouts are sampled at temperature $1.0$ to provide within-group variation, whereas deployment uses greedy decoding.
\begin{table}[h!]\small
    \caption{\textsc{MemOpt} training hyperparameters.}
    \label{tab:memopt_hyperparameters}
    \begin{center}
    \begin{tabular}{llc}
        \toprule
        Group & Parameter & Value \\
        \midrule
        \multirow{4}{*}{Sampling}
          & Group size $G$ (rollouts per segment) & 5 \\
          & Rollout temperature / top-$p$ / top-$k$ & 1.0 / 1.0 / disabled \\
          & Maximum prompt length (tokens) & 12{,}288 \\
          & Maximum response length (tokens) & 512 \\
        \midrule
        \multirow{4}{*}{Optimization}
          & Learning rate & $1\times10^{-6}$ \\
          & Weight decay & 0.01 \\
          & Learning-rate warmup & none \\
          & Gradient-norm clip & 1.0 \\
        \midrule
        \multirow{4}{*}{Objective}
          & Policy clipping ratio $\epsilon_{\mathrm p}$ (symmetric) & 0.2 \\
          & Advantage clip $\kappa$ & 3.0 \\
          & KL penalty $\beta$ (loss term) & 0.01 \\
          & Entropy coefficient & 0 \\
        \midrule
        \multirow{4}{*}{Schedule}
          & Segments per optimizer step & 32 \\
          & Mini-batch size & 16 \\
          & Inner epochs per step & 1 \\
          & Training epochs & 3 \\
        \bottomrule
    \end{tabular}
    \end{center}
\end{table}

\paragraph{Training data.} Of the 4,445 segments in the training histories, 1,451 (32.6\%) provide verified evidence for at least one training question and are optimized directly. The other segments are not sampled for writer RL but remain in the memory bank, which is rebuilt with the current writer after each epoch (Appendix~\ref{app:retrievability_replay}), so retrievability is measured against all entries. Writer rewards use only the 1,653 training questions that have at least one verified evidence segment, forming 3,055 question--segment pairs. Table~\ref{tab:rev_category_supervision} reports their distribution over the SuperMemory-VQA categories. We do not penalize content unrelated to the training questions, because supported details may be needed by future questions. Faithfulness instead penalizes unsupported content, and the KL term keeps the writer close to its initialization. Because the rewards require questions linked to evidence segments, adapting \textsc{MemOpt} to a new domain requires such examples, which can be generated from recorded lifelogs as done for SuperMemory-VQA.

\begin{table}[h!]\small
    \caption{Distribution of \textsc{MemOpt} training supervision over SuperMemory-VQA question categories. Counts include only training questions with at least one verified evidence segment.}
    \label{tab:rev_category_supervision}
    \begin{center}
    \resizebox{\linewidth}{!}{
    \begin{tabular}{lccccccc}
        \toprule
        & \makecell{Conversational\\Memory} & \makecell{Intent\\Recall} & \makecell{In-Context\\Retrieval}
        & \makecell{Timeline\\Reconstruction} & \makecell{Object-Location\\Memory} & \makecell{Visual\\Recall} & Total \\
        \midrule
        Training questions & 396 & 238 & 311 & 317 & 209 & 182 & 1,653 \\
        Question--segment pairs & 602 & 469 & 709 & 762 & 277 & 236 & 3,055 \\
        \bottomrule
    \end{tabular}
    }
    \end{center}
\end{table}

\paragraph{Compute.} We train the writer on two nodes with eight H100 80GB GPUs each. One node runs GRPO with verl and vLLM rollouts, and the other serves the frozen Qwen3.5-9B evaluators as four replicas of two GPUs each. Each 45-step epoch takes about 6 hours, and a full three-epoch run, including checkpoint selection, takes about 25 hours and about 360 H100 GPU-hours, as some steps run on a single node. The one-time preparation before training, including key-fact extraction and the first reader-trace collection, takes about 20 additional GPU-hours.

\subsection{Retrievability Trace Collection and Replay}
\label{app:retrievability_replay}

At the beginning of each training epoch, we build a memory bank with the current writer and run the frozen agentic reader on every training question associated with at least one verified evidence segment. For each question, we cache the search queries, time intervals, retrieval budgets, and fetched intervals issued before the final answer. These traces are reused to score all candidate memories sampled during that epoch.

To evaluate a candidate $y$ for segment $c$, we replace only the corresponding entry in the memory bank and replay every cached action associated with questions in $\mathcal{Q}(c)$. For a search action, we recompute the candidate's similarity and rank it against all entries eligible under that action's causal and temporal constraints. The action returns $y$ only when its rank falls within the recorded retrieval budget. For a fetch action, $y$ is returned when the requested interval contains $c$. This replay determines membership in $\mathcal{C}_q(y)$ without executing a new agent reasoning trajectory. The memory bank and traces are rebuilt after each epoch.

\subsection{Token-Level Group-Relative Objective}
\label{app:memopt_optimization}

Equations~\ref{eq:memopt_group_statistics} and \ref{eq:memopt_token_advantage} define the token advantages used for optimization. We clip each $\widetilde{A}_{i,t}$ to $[-\kappa,\kappa]$, then recenter and whiten the values over generated response tokens while excluding prompt and padding positions. We denote the resulting advantage by $A_{i,t}$. Let $\pi_{\theta_{\mathrm{old}}}$ denote the policy that sampled the current candidates. Its token probability ratio with the updated writer is
\begin{equation}
    \rho_{i,t}(\theta)
    =\frac{\pi_\theta(y_{i,t}\mid c,y_{i,<t})}
    {\pi_{\theta_{\mathrm{old}}}(y_{i,t}\mid c,y_{i,<t})}.
    \label{eq:memopt_policy_ratio}
\end{equation}
The \textsc{MemOpt} policy objective is
\begin{equation}
\begin{aligned}
    \mathcal{L}_{\textsc{MemOpt}}(\theta)
    =-\mathbb{E}_{c,i,t}\Big[&
    \min\!\big(
        \rho_{i,t}(\theta)A_{i,t},
        \operatorname{clip}(\rho_{i,t}(\theta),1-\epsilon_{\mathrm p},1+\epsilon_{\mathrm p})A_{i,t}
    \big)
    -\beta\widehat{D}^{\mathrm{KL}}_{i,t}\Big],
\end{aligned}
\label{eq:memopt_objective}
\end{equation}
where $\epsilon_{\mathrm p}$ is the policy clipping ratio, $\widehat{D}^{\mathrm{KL}}_{i,t}$ is the per-token Kullback--Leibler estimate between $\pi_\theta$ and the frozen initial writer $\pi_{\mathrm{ref}}$, and $\beta$ controls its strength. Table~\ref{tab:memopt_hyperparameters} lists the numerical settings.

\section{Controlled Comparison with Trained Baselines}
\label{app:controlled_trained_baselines}

The released checkpoints in the main comparison preserve the original systems but leave differences in model scale and training data. We therefore conduct an additional controlled comparison in which VST, TaskMem, and M3-Agent are all reproduced with Qwen3.5-9B and trained only on the SuperMemory-VQA training split used by \textsc{MemOpt}. For reference, Table~\ref{tab:controlled_trained_baselines} also lists the released checkpoints of all three methods.

\begin{table}[h!]\small
    \caption{Controlled comparison with trained baselines. Reproduced methods and \textsc{MemLife} + \textsc{MemOpt} use Qwen3.5-9B and task-specific training data only from SuperMemory-VQA, while released checkpoints keep their original backbones and training data. EgoLifeQA is evaluated out of distribution, and bold marks the best result.}
    \label{tab:controlled_trained_baselines}
    \begin{center}
    \begin{tabular}{lcccc}
        \toprule
        \multirow{2}{*}{Method}
        & \multicolumn{2}{c}{SuperMemory-VQA}
        & \multicolumn{2}{c}{EgoLifeQA} \\
        \cmidrule(lr){2-3}\cmidrule(l){4-5}
        & Accuracy & Recall & Accuracy & Recall \\
        \midrule
        VST (released) & 37.56 & 56.87 & 33.60 & 3.40 \\
        VST (reproduced) & 56.50 & 65.08 & 29.40 & 7.60 \\
        TaskMem (released) & 50.88 & 65.08 & 46.20 & 40.20 \\
        TaskMem (reproduced) & 48.96 & 51.15 & 48.20 & 35.00 \\
        M3-Agent (released) & 43.18 & 31.87 & 30.40 & 7.80 \\
        M3-Agent (reproduced) & 53.93 & 54.20 & 35.40 & 9.60 \\
        \textsc{MemLife} + \textsc{MemOpt} & \textbf{60.35} & \textbf{81.49} & \textbf{56.60} & \textbf{50.60} \\
        \bottomrule
    \end{tabular}
    \end{center}
\end{table}

Table~\ref{tab:controlled_trained_baselines} shows that \textsc{MemLife} with \textsc{MemOpt} achieves the highest accuracy and recall on both the in-domain SuperMemory-VQA test set and out-of-domain EgoLifeQA. Across the three baselines, neither the released nor the reproduced version is consistently better. The reproductions and \textsc{MemOpt} use no EgoLifeQA data for task-specific training. Moreover, \textsc{MemOpt} updates only the memory writer, whereas VST and M3-Agent also adapt their answering or control components. Together with the primary comparison in Table~\ref{tab:main_results}, these results indicate that the gains from \textsc{MemOpt} are not explained by backbone scale or differences in task-specific training data.

\section{Stability and Robustness Analysis}
\label{app:stability}
\paragraph{Decoding stability.} In the main experiments, we use deterministic decoding across all methods for fairness and reproducibility.
To assess stability under stochastic decoding, we repeat inference five times at temperature 1.0 for \textsc{MemLife}, \textsc{MemLife-V}, and EgoRAG, the competing system with the highest average accuracy. Table~\ref{tab:decoding_stability} compares the greedy results from the main evaluation with the mean and standard deviation over five sampled runs.

\begin{table}[h!] \small
    \caption{Stability across reader decoding settings. $T=0.0$ reports greedy decoding, while $T=1.0$ reports mean $\pm$ standard deviation over five runs.}
    \label{tab:decoding_stability}
    \begin{center}
    \resizebox{\linewidth}{!}{
    \begin{tabular}{lcccccccc}
        \toprule
        \multirow{4}{*}{Method}
        & \multicolumn{4}{c}{SuperMemory-VQA}
        & \multicolumn{4}{c}{EgoLifeQA} \\
        \cmidrule(lr){2-5}\cmidrule(l){6-9}
        & \multicolumn{2}{c}{Accuracy} & \multicolumn{2}{c}{Recall}
        & \multicolumn{2}{c}{Accuracy} & \multicolumn{2}{c}{Recall} \\
        \cmidrule(lr){2-3}\cmidrule(lr){4-5}\cmidrule(lr){6-7}\cmidrule(l){8-9}
        & $T=0.0$ & $T=1.0$ & $T=0.0$ & $T=1.0$
        & $T=0.0$ & $T=1.0$ & $T=0.0$ & $T=1.0$ \\
        \midrule
        EgoRAG            & 49.28 & 49.18 $\pm$ 0.95 & 66.79 & 66.79 $\pm$ 0.00 & 48.20 & 46.48 $\pm$ 0.95 & 29.60 & 28.80 $\pm$ 0.00 \\
        \textsc{MemLife}   & 56.50 & 55.31 $\pm$ 1.53 & 80.53 & 80.69 $\pm$ 0.32 & 52.80 & 50.16 $\pm$ 1.83 & 48.40 & 47.48 $\pm$ 1.36 \\
        \textsc{MemLife}-V & 57.78 & 58.78 $\pm$ 0.41 & 84.35 & 84.58 $\pm$ 0.64 & 53.20 & 52.44 $\pm$ 1.72 & 50.20 & 49.72 $\pm$ 0.30 \\
        \midrule
        \multirow{4}{*}{Method}
        & \multicolumn{4}{c}{SuperMemory-LVQA}
        & \multicolumn{4}{c}{EgoLife-EQA} \\
        \cmidrule(lr){2-5}\cmidrule(l){6-9}
        & \multicolumn{2}{c}{Accuracy} & \multicolumn{2}{c}{Recall}
        & \multicolumn{2}{c}{Accuracy} & \multicolumn{2}{c}{Recall} \\
        \cmidrule(lr){2-3}\cmidrule(lr){4-5}\cmidrule(lr){6-7}\cmidrule(l){8-9}
        & $T=0.0$ & $T=1.0$ & $T=0.0$ & $T=1.0$
        & $T=0.0$ & $T=1.0$ & $T=0.0$ & $T=1.0$ \\
        \midrule
        EgoRAG            & 47.51 & 48.02 $\pm$ 0.43 & 47.90 & 48.85 $\pm$ 0.00 & 38.00 & 36.20 $\pm$ 2.39 & 13.00 & 14.00 $\pm$ 0.00 \\
        \textsc{MemLife}   & 56.18 & 55.31 $\pm$ 0.91 & 46.18 & 45.00 $\pm$ 0.97 & 52.00 & 50.80 $\pm$ 4.66 & 44.00 & 42.20 $\pm$ 2.95 \\
        \textsc{MemLife}-V & 57.95 & 58.17 $\pm$ 1.24 & 48.47 & 48.51 $\pm$ 1.93 & 50.00 & 53.20 $\pm$ 3.27 & 45.00 & 45.00 $\pm$ 0.71 \\
        \bottomrule
    \end{tabular}
    }
    \end{center}
\end{table}

The performance ordering remains stable under sampled decoding. \textsc{MemLife} and \textsc{MemLife}-V outperform EgoRAG in accuracy on every benchmark and generally provide higher recall. The gains remain consistent across runs, and the accuracy improvements of \textsc{MemLife} over EgoRAG are significant on all four benchmarks ($p<0.05$). The recall of EgoRAG has a zero standard deviation, since it has a fixed retrieval process instead of using agentic search as \textsc{MemLife}.

\paragraph{Backbones from other model families.} Table~\ref{tab:other_backbones} evaluates \textsc{MemLife}, EgoRAG, and EGAgent with two backbones from other model families, the open-weight Gemma4-31B and the proprietary Gemini3.5-Flash. EgoRAG is the strongest training-free baseline in Table~\ref{tab:main_results}, and EGAgent shares the most designs with \textsc{MemLife} (Table~\ref{tab:rev_design_comparison}). Within each backbone, all replaceable model components of every system use that backbone, and each system otherwise keeps its configuration from Table~\ref{tab:main_results}. \textsc{MemLife} outperforms both baselines on both benchmarks with every backbone.

\begin{table}[h!]\small
    \caption{Accuracy of \textsc{MemLife} and representative training-free baselines with backbones from different model families.}
    \label{tab:other_backbones}
    \begin{center}
    \begin{tabular}{llcc}
        \toprule
        Backbone & Method & SuperMemory-VQA & EgoLifeQA \\
        \midrule
        \multirow{3}{*}{Qwen3.5-9B} & EgoRAG & 49.28 & 48.20 \\
         & EGAgent & 43.66 & 34.00 \\
         & \textsc{MemLife} & 56.50 & 52.80 \\
        \midrule
        \multirow{3}{*}{Gemma4-31B} & EgoRAG & 49.44 & 51.40 \\
         & EGAgent & 53.61 & 44.60 \\
         & \textsc{MemLife} & 67.90 & 67.40 \\
        \midrule
        \multirow{3}{*}{Gemini3.5-Flash} & EgoRAG & 52.65 & 55.80 \\
         & EGAgent & 61.64 & 46.80 \\
         & \textsc{MemLife} & 68.38 & 69.20 \\
        \bottomrule
    \end{tabular}
    \end{center}
\end{table}

\paragraph{Significance of \textsc{MemOpt} with the 27B reader.} With the Qwen3.6-27B reader, the greedy gains of \textsc{MemOpt} in Table~\ref{tab:writer_reader_backbones} are small. To test their significance, we repeat question answering with this reader five times at temperature 1.0, using memories written by the default Qwen3.5-9B writer before and after \textsc{MemOpt}, as well as oracle context. We average the correctness of each question across the five runs and obtain 95\% confidence intervals and $p$-values from paired bootstrap and permutation tests over questions.

\begin{table}[h!]\small
    \caption{Five runs with the Qwen3.6-27B reader at temperature 1.0. Accuracy (\%) is the mean $\pm$ standard deviation across runs. Differences report 95\% confidence intervals and $p$-values from paired tests at the question level.}
    \label{tab:rev_27b_significance}
    \begin{center}
    \begin{tabular}{lcc}
        \toprule
        Condition & SuperMemory-VQA & EgoLifeQA \\
        \midrule
        Zero-shot memory & 66.65 $\pm$ 0.84 & 57.40 $\pm$ 1.39 \\
        \textsc{MemOpt} memory & 65.14 $\pm$ 1.05 & 61.56 $\pm$ 2.48 \\
        Oracle context & 68.22 $\pm$ 0.49 & 68.68 $\pm$ 0.95 \\
        \midrule
        \textsc{MemOpt} $-$ zero-shot & $-1.51$ ([$-3.95$, $+0.87$], $p=0.24$) & $+4.16$ ([$+1.36$, $+6.96$], $p=0.004$) \\
        Oracle $-$ zero-shot & $+1.57$ ([$-1.96$, $+5.04$], $p=0.39$) & $+11.28$ ([$+7.00$, $+15.60$], $p<0.001$) \\
        \bottomrule
    \end{tabular}
    \end{center}
\end{table}

Table~\ref{tab:rev_27b_significance} shows that the gain of \textsc{MemOpt} depends on the reader's headroom. On EgoLifeQA, where zero-shot memory remains far below oracle context, \textsc{MemOpt} significantly improves accuracy ($p<0.01$), showing that the optimized memory also benefits a reader that was not used to collect retrievability supervision. On SuperMemory-VQA, zero-shot memory already brings the 27B reader close to oracle context, leaving little room for any memory to help, and the change from \textsc{MemOpt} is not significant. With the 9B reader, which leaves more headroom, \textsc{MemOpt} improves accuracy on all four benchmarks (Table~\ref{tab:main_results}).

\paragraph{Subject-balanced training.} The training subjects contribute uneven numbers of questions. To test the sensitivity of \textsc{MemOpt} to this composition, we retrain the writer on a subject-balanced subset in which each of S1--S6 contributes 177 questions, the number available for the smallest subject, S3. Table~\ref{tab:rev_subject_balanced} shows that \textsc{MemOpt} still improves \textsc{MemLife} on both benchmarks and retains most of its gains despite using 36\% fewer training questions. The benefit of writer optimization therefore does not rely on the uneven subject distribution and persists with fewer training questions.

\begin{table}[h!]\small
    \caption{Subject-balanced \textsc{MemOpt} training. Counts include only training questions with at least one verified evidence segment.}
    \label{tab:rev_subject_balanced}
    \begin{center}
    \resizebox{\linewidth}{!}{
    \begin{tabular}{lccccccccccc}
        \toprule
        \multirow{2.5}{*}{Training set}
        & \multicolumn{7}{c}{Training questions}
        & \multicolumn{2}{c}{SuperMemory-VQA}
        & \multicolumn{2}{c}{EgoLifeQA} \\
        \cmidrule(lr){2-8}\cmidrule(lr){9-10}\cmidrule(l){11-12}
        & S1 & S2 & S3 & S4 & S5 & S6 & Total & Accuracy & Recall & Accuracy & Recall \\
        \midrule
        None (\textsc{MemLife}) & -- & -- & -- & -- & -- & -- & -- & 56.50 & 80.53 & 52.80 & 48.40 \\
        Original & 475 & 395 & 177 & 188 & 204 & 214 & 1,653 & 60.35 & 81.49 & 56.60 & 50.60 \\
        Subject-balanced & 177 & 177 & 177 & 177 & 177 & 177 & 1,062 & 59.07 & 81.30 & 55.40 & 49.20 \\
        \bottomrule
    \end{tabular}
    }
    \end{center}
\end{table}

\section{Efficiency and Retrieval Behavior Analysis}
\label{app:efficiency}

\paragraph{Storage and reading time.} Table~\ref{tab:efficiency} compares storage and question-time costs across various systems, which are the key efficiency concerns in runtime question answering. Storage is measured over matched source histories and normalized by video duration, while reading time excludes model initialization. 

\begin{table}[h!]\small
    \caption{Storage and reading efficiency of training-free video memory systems. Storage is normalized by source-video duration, and reading time is averaged per question. Dashes indicate methods without a persistent memory store.}
    \label{tab:efficiency}
    \begin{center}
    \begin{tabular}{lcccc}
        \toprule
        \multirow{2.5}{*}{Method}
        & \multicolumn{2}{c}{Memory Storage (MB/h)}
        & \multicolumn{2}{c}{Reading Speed (s/question)} \\
        \cmidrule(lr){2-3}\cmidrule(l){4-5}
        & SuperMemory-VQA & EgoLifeQA
        & SuperMemory-VQA & EgoLifeQA \\
        \midrule
        Video ReCap        & 3.86  & 3.66  & \underline{20.4} & 54.2 \\
        EgoRAG             & 4.98  & 5.49  & 33.2 & \underline{43.5} \\
        Video-RAG          & \underline{1.03}  & \underline{3.44}  & 60.9 & 76.9 \\
        VideoARM           & --  & --  & 47.5 & 110.4 \\
        EGAgent            & 21.50 & 22.20 & 200.3 & 230.7 \\
        WorldMM            & 2.09  & 3.88  & 44.7 & 57.1 \\
        \textsc{MemLife}   & \textbf{0.68} & \textbf{0.67} & \textbf{13.1} & \textbf{20.7} \\
        \bottomrule
    \end{tabular}
    \end{center}
\end{table}

\textsc{MemLife} has the smallest footprint among systems with persistent storage, which has nearly unchanged storage per video hour across the two benchmarks. It also provides the fastest reading time. Its informative text episodes and targeted retrieval typically expose sufficient evidence within a few agent rounds, while chronological ordering reduces the subsequent reasoning burden. The default reader also avoids processing source-video tokens at question time. Together, these choices limit unnecessary generation and multimodal processing, helping explain the lower latency relative to baselines with different execution patterns.

\paragraph{Input tokens.} Table~\ref{tab:rev_tokens} reports the mean input tokens per question. Total input tokens sum the input across all reader calls, including history repeated in later rounds, while new input tokens count only the input added in each round. \textsc{MemLife} uses more total input tokens than the single-pass EgoRAG, because its multi-round reader receives the earlier history again in each call, but a comparable number of new input tokens, while achieving higher accuracy. When each search returns $k=8$ instead of 32 entries, both token counts decrease, \textsc{MemLife} still outperforms all training-free baselines on both benchmarks, and \textsc{MemOpt} further improves it. On SuperMemory-VQA, \textsc{MemLife} + \textsc{MemOpt} at $k=8$ matches the accuracy of \textsc{MemLife} at $k=32$ with less than half of the total input tokens.

\begin{table}[h!]\small
    \caption{Accuracy (\%) and mean input tokens per question. Total counts all reader input across rounds, New counts only the input added in each round, and $k$ is the number of entries returned per search.}
    \label{tab:rev_tokens}
    \begin{center}
    \begin{tabular}{lcccccc}
        \toprule
        \multirow{2.5}{*}{Method}
        & \multicolumn{3}{c}{SuperMemory-VQA}
        & \multicolumn{3}{c}{EgoLifeQA} \\
        \cmidrule(lr){2-4}\cmidrule(l){5-7}
        & Accuracy & Total & New & Accuracy & Total & New \\
        \midrule
        Video ReCap & 36.28 & 10k & 10k & 35.20 & 53k & 53k \\
        EgoRAG & 49.28 & 24k & 24k & 48.20 & 24k & 24k \\
        Video-RAG & 43.98 & 18k & 18k & 38.80 & 19k & 19k \\
        VideoARM & 39.33 & 59k & 46k & 36.00 & 183k & 118k \\
        EGAgent & 43.66 & 17k & 15k & 34.00 & 29k & 26k \\
        WorldMM & 42.05 & 31k & 24k & 42.20 & 46k & 37k \\
        \midrule
        \textsc{MemLife} ($k=32$) & 56.50 & 58k & 24k & 52.80 & 45k & 20k \\
        \textsc{MemLife} + \textsc{MemOpt} ($k=32$) & 60.35 & 70k & 28k & 56.60 & 77k & 32k \\
        \textsc{MemLife} ($k=8$) & 53.93 & 36k & 10k & 51.00 & 37k & 11k \\
        \textsc{MemLife} + \textsc{MemOpt} ($k=8$) & 56.34 & 28k & 9k & 52.60 & 45k & 14k \\
        \bottomrule
    \end{tabular}
    \end{center}
\end{table}

\paragraph{Memory length.} Table~\ref{tab:rev_memory_length} reports the average memory length before and after \textsc{MemOpt} under the same 512-token generation cap (Appendix~\ref{app:inference_details}). \textsc{MemOpt} writes longer memories, especially on EgoLifeQA. During training, the average response length is non-monotonic and remains well below this cap (Figure~\ref{fig:memopt_training_dynamics}), suggesting that the rewards do not simply favor longer memories.

\begin{table}[h!]\small
    \caption{Average memory length in tokens before and after \textsc{MemOpt}.}
    \label{tab:rev_memory_length}
    \begin{center}
    \begin{tabular}{lccc}
        \toprule
        Benchmark & \# Entries & Zero-shot & \textsc{MemOpt} \\
        \midrule
        SuperMemory-VQA & 752 & 245.5 & 280.6 \\
        EgoLifeQA & 6,250 & 237.7 & 370.6 \\
        \bottomrule
    \end{tabular}
    \end{center}
\end{table}

\paragraph{Retrieval behavior.} Table~\ref{tab:rev_rounds} reports the average number of rounds and tool calls per question. Each round contains either one tool call or the final answer, so a two-round trajectory contains one retrieval call followed by the answer. Although ten rounds are available, the reader uses about three rounds per question. Most retrieval calls are \textsc{SearchMemory}, which already supports temporal constraints through its optional time interval. On EgoLifeQA, about 9--11\% of searches use this interval, for example by resolving ``today'' in a question to the corresponding day, whereas it is almost never used on SuperMemory-VQA. \textsc{FetchMemory} is called in only 1--7 questions per benchmark to inspect consecutive episodes within a selected interval. With the same reader, \textsc{MemOpt} improves accuracy while reducing the number of retrieval calls on both benchmarks, especially on EgoLifeQA, indicating that the optimized memories improve retrieval efficiency as well as effectiveness. Table~\ref{tab:rev_round_groups} further groups questions by the number of rounds used. With \textsc{MemOpt}, more questions are answered within two rounds, accuracy in this group improves on both benchmarks, and accuracy also increases in most other groups. Appendix~\ref{app:rev_trajectories} presents representative trajectories.

\begin{table}[h!]\small
    \caption{Rounds and tool calls per question with the same Qwen3.5-9B reader. Each round contains one tool call or the final answer.}
    \label{tab:rev_rounds}
    \begin{center}
    \resizebox{\linewidth}{!}{
    \begin{tabular}{llccccc}
        \toprule
        Benchmark & Memory & Accuracy & Recall & Rounds / q & \textsc{SearchMemory} / q & \textsc{FetchMemory} / q \\
        \midrule
        \multirow{2}{*}{SuperMemory-VQA} & \textsc{MemLife} & 56.50 & 80.53 & 3.39 & 2.39 & 0.008 \\
        & + \textsc{MemOpt} & 60.35 & 81.49 & 3.32 & 2.32 & 0.002 \\
        \midrule
        \multirow{2}{*}{EgoLifeQA} & \textsc{MemLife} & 52.80 & 48.40 & 3.32 & 2.30 & 0.014 \\
        & + \textsc{MemOpt} & 56.60 & 50.60 & 2.99 & 1.98 & 0.008 \\
        \bottomrule
    \end{tabular}
    }
    \end{center}
\end{table}

\begin{table}[h!]\small
    \caption{Accuracy (\%) grouped by the number of rounds used, with the number of questions in parentheses.}
    \label{tab:rev_round_groups}
    \begin{center}
    \begin{tabular}{llcccc}
        \toprule
        Benchmark & Memory & 2 rounds & 3--5 rounds & 6--9 rounds & 10 rounds \\
        \midrule
        \multirow{2}{*}{SuperMemory-VQA} & \textsc{MemLife} & 63.0 (281) & 52.5 (265) & 47.1 (68) & 44.4 (9) \\
        & + \textsc{MemOpt} & 65.7 (309) & 58.2 (239) & 41.7 (60) & 60.0 (15) \\
        \midrule
        \multirow{2}{*}{EgoLifeQA} & \textsc{MemLife} & 59.3 (290) & 46.3 (136) & 39.6 (53) & 38.1 (21) \\
        & + \textsc{MemOpt} & 61.5 (296) & 48.8 (160) & 51.4 (37) & 57.1 (7) \\
        \bottomrule
    \end{tabular}
    \end{center}
\end{table}

\section{Ablation of Retrievability Approximation}

Computing retrievability requires specifying how a future question will access a candidate memory. We compare two approximations. The simplified variant uses the original question directly as a fixed semantic-search query and scores whether the candidate is retrieved. It therefore requires neither a reader policy nor a reader rollout during writer training. Standard \textsc{MemOpt} instead runs the \textsc{MemLife} agentic reader at the beginning of each epoch and collects its memory-access actions. These actions reflect the agent's reformulated queries and time-scoped retrieval decisions, but remain fixed within the epoch to avoid candidate-dependent variation. After writer training, we evaluate both variants with the standard \textsc{MemLife} agentic reader, which performs multi-round memory access and chronologically orders the retrieved entries before answering.

\begin{table}[h!]\small
    \caption{Effect of the training-time retrievability approximation on answer accuracy (\%).}
    \label{tab:retrievability_approximation}
    \begin{center}
    \begin{tabular}{lcc}
        \toprule
        Training-time proxy & SuperMemory-VQA & EgoLifeQA \\
        \midrule
        None (zero-shot) & 56.50 & 52.80 \\
        Original question & 57.62 & 55.40 \\
        Agent actions & 60.35 & 56.60 \\
        \bottomrule
    \end{tabular}
    \end{center}
\end{table}

Both approximations outperform the zero-shot writer on both benchmarks. The original-question proxy already provides useful supervision without requiring a training-time reader rollout. Agent actions perform best on both datasets, indicating that reformulated queries and time-scoped accesses expose retrieval failures that the original question alone may miss. We therefore use agent actions in the standard \textsc{MemOpt} configuration, while retaining the original-question proxy as a simpler, reader-policy-independent alternative.

\section{Reward Dynamics during \textsc{MemOpt} Training}
\label{app:reward_dynamics}

\begin{figure}[h!]
    \centering
    \includegraphics[width=1\linewidth]{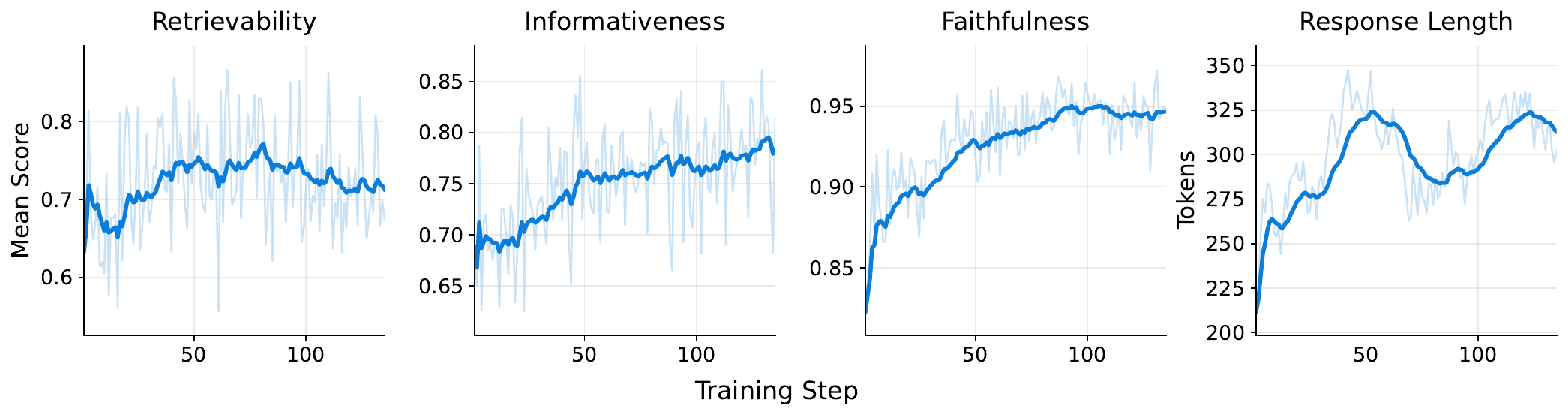}
    \caption{Dynamics of the three reward signals and the average response length during \textsc{MemOpt} training.}
    \label{fig:memopt_training_dynamics}
\end{figure}

To examine how the three reward signals evolve during optimization, we record their scores over three epochs, comprising 135 training steps. Retrievability and informativeness are averaged across the sampled candidate memories. Since faithfulness is evaluated at the token level, we first take the minimum token score within each candidate and then average these minima across candidates. This conservative aggregation prevents a few unsupported tokens from being obscured by many well-supported ones. In Figure~\ref{fig:memopt_training_dynamics}, the light curves show the resulting per-step scores, while the bold curves show their smoothed trends.

Following an initial fluctuation, retrievability improves mainly during the early stages of training and stabilizes sooner, although its per-step values remain noisy because they depend on each candidate's position within the evolving memory store and the cached reader actions. Informativeness and faithfulness improve for longer, indicating that the writer continues to preserve more answer-relevant evidence and produce better-grounded memories. Together, these trends suggest that \textsc{MemOpt} improves all three dimensions without a sustained trade-off in retrievability. Figure~\ref{fig:memopt_training_dynamics} also shows the average response length, which rises in the early steps, decreases in the middle of training, and rises again later, while remaining well below the 512-token cap. This non-monotonic trend suggests that the rewards do not simply favor longer memories.

\section{Discussion on Memory Reader Training}
\label{app:reader_training}

Although \textsc{MemOpt} focuses on the writer, the agentic reader can also be trained to improve memory access and answer generation. We compare training neither component, the writer alone, the reader alone, and both components. For reader reinforcement learning, the reward is $(\text{accuracy}+\text{recall})/2$. To reduce training cost, we cap trajectories at six interaction rounds, compared with ten rounds at evaluation. A trajectory containing a formatting error or exhausting this budget will receive a reward of $-1$. We introduced the budget penalty because a reader trained with the combined reward but without this penalty always used all ten rounds at evaluation. The six-round cap is rarely restrictive, as the untrained reader finishes within six rounds on 93.9\% of SuperMemory-VQA and 89.8\% of EgoLifeQA questions when ten rounds are available.

\begin{table}[h!]\small
    \caption{Effects of writer and reader training. All values are percentages.}
    \label{tab:reader_training}
    \begin{center}
    \begin{tabular}{cccccc}
        \toprule
        \multirow{2}{*}{Writer trained} & \multirow{2}{*}{Reader trained}
        & \multicolumn{2}{c}{SuperMemory-VQA}
        & \multicolumn{2}{c}{EgoLifeQA} \\
        \cmidrule(lr){3-4}\cmidrule(l){5-6}
        & & Accuracy & Recall & Accuracy & Recall \\
        \midrule
        \redcross   & \redcross   & 56.50 & 80.53 & 52.80 & 48.40 \\
        \greencheck & \redcross   & 60.35 & 81.49 & 56.60 & 50.60 \\
        \redcross   & \greencheck & 61.00 & 89.50 & 51.40 & 59.00 \\
        \greencheck & \greencheck & 63.08 & 87.98 & 54.20 & 56.80 \\
        \bottomrule
    \end{tabular}
    \end{center}
\end{table}

Reader training substantially improves both accuracy and recall on the in-domain SuperMemory-VQA benchmark. Adding writer training to the trained reader further improves accuracy with only a small decrease in recall, indicating that the optimized memories preserve more answer-useful information rather than merely maximizing evidence retrieval. On the out-of-domain EgoLifeQA benchmark, reader training still improves recall but does not consistently improve answer accuracy, whereas writer-only training achieves the highest accuracy. This contrast suggests that reader policies more readily specialize to the training question and source-video distributions, as well as the resulting interaction trajectories. Writer training instead improves the persistent memory itself, allowing different readers and evaluation settings to benefit from the same higher-quality memory or writing policy. We therefore center \textsc{MemOpt} on writer optimization. Reader training provides complementary in-domain gains, while making these gains consistent under distribution shift remains future work.

To test whether the EgoLifeQA result stems from rewarding recall, we also train a reader with an accuracy-only reward, keeping the six-round budget and all other settings. Table~\ref{tab:rev_reader_rewards} compares the two reader rewards with writer-only training. The recall term does encourage broader retrieval, as the reader trained with $(\text{accuracy}+\text{recall})/2$ uses more rounds than the untrained reader and attains higher recall. However, removing the recall term does not restore EgoLifeQA accuracy. The accuracy-only reader converges to answering after a single search, which further improves in-domain accuracy but lowers EgoLifeQA accuracy. One possible explanation is that a single search often suffices on the short SuperMemory-VQA histories but not on the much longer EgoLifeQA history (Table~\ref{tab:dataset_statistics}), where the recall of this reader drops. Under both rewards, reader training improves in-domain accuracy without transferring to EgoLifeQA, whereas writer-only \textsc{MemOpt} improves EgoLifeQA accuracy with the same untrained reader.

\begin{table}[h!]\small
    \caption{Reader rewards compared with writer-only training. Rounds count all reader calls per question, including the final answer.}
    \label{tab:rev_reader_rewards}
    \begin{center}
    \begin{tabular}{lcccccc}
        \toprule
        \multirow{2.5}{*}{Training}
        & \multicolumn{3}{c}{SuperMemory-VQA}
        & \multicolumn{3}{c}{EgoLifeQA} \\
        \cmidrule(lr){2-4}\cmidrule(l){5-7}
        & Accuracy & Recall & Rounds & Accuracy & Recall & Rounds \\
        \midrule
        None & 56.50 & 80.53 & 3.39 & 52.80 & 48.40 & 3.32 \\
        Reader RL, $(\text{accuracy}+\text{recall})/2$ & 61.00 & 89.50 & 4.79 & 51.40 & 59.00 & 3.83 \\
        Reader RL, accuracy only & 66.13 & 73.66 & 2.00 & 49.80 & 42.80 & 2.00 \\
        Writer RL, \textsc{MemOpt} & 60.35 & 81.49 & 3.32 & 56.60 & 50.60 & 2.99 \\
        \bottomrule
    \end{tabular}
    \end{center}
\end{table}

\section{Predecessor Context for Memory Writing}
\label{app:predecessor_context}

\textsc{MemLife} writes each video segment independently. To examine whether continuity across adjacent segments benefits memory construction, we compare this design with two alternatives that condition the writer on the immediately preceding segment. The text variant provides both the previous transcript and memory entry, while the multimodal variant additionally provides the previous frames. The independent variant (None) omits the predecessor block and uses the multimodal information from the corresponding segment as the only context. 

\begin{table}[h!]\small
    \caption{Answer accuracy across SuperMemory-VQA question types under different forms of predecessor context for memory writing. Bold values indicate the best result in each column.}
    \label{tab:predecessor_context}
    \begin{center}
    \resizebox{\linewidth}{!}{
    \begin{tabular}{lccccccc}
        \toprule
        \makecell[l]{Predecessor\\Context}
        & \makecell{Conversational\\Memory}
        & \makecell{Intent\\Recall}
        & \makecell{In-Context\\Retrieval}
        & \makecell{Timeline\\Reconstruction}
        & \makecell{Object-Location\\Memory}
        & \makecell{Visual\\Recall}
        & Overall \\
        \midrule
        None & 63.41 & 77.42 & \textbf{43.18} & \textbf{57.41} & \textbf{52.29} & 44.12 & \textbf{56.50} \\
        {Text} & 62.60 & \textbf{86.02} & 37.50 & 48.15 & 45.87 & \textbf{45.10} & 54.25 \\
        {Multimodal} & \textbf{69.92} & 82.80 & 37.50 & 47.22 & 46.79 & 43.14 & 54.90 \\
        \bottomrule
    \end{tabular}
    }
    \end{center}
\end{table}

Table \ref{tab:predecessor_context} shows that predecessor context produces mixed effects. Text context improves intent and visual recall, while multimodal context improves conversational and intent recall. Both variants reduce performance on in-context retrieval, timeline reconstruction, and object-location memory, and neither improves overall accuracy. We therefore retain independent segment writing, which also avoids sequential dependencies and error propagation during memory construction.

\section{Memory from Text and Multimodal Inputs}
\label{app:text_memory}

Text-input agent memory systems maintain persistent information from language-based histories, such as conversations, documents, or natural-language observations. They determine what information to store and how to update, organize, and retrieve it for later use. Existing systems store observations and synthesize reflections \citep{park2023generative}, manage tiered memory stores \citep{packer2023memgpt,kang2025memoryos}, or maintain and consolidate information from conversations \citep{zhong2024memorybank,chhikara2025mem0}. Other architectures organize histories through segmentation, compression, or dynamically linked and graph-structured memories \citep{pan2025secom,xu2025amem,wang2026sage}.

Beyond architectural design, recent methods learn what to retain or how to manage memory. They optimize memory operations or construction from downstream feedback \citep{yan2025memoryr1,wang2025mema}, retain source evidence under a fixed budget \citep{li2026ember}, or provide denser supervision for different stages and components of memory construction and use \citep{shen2026membuilder,zhang2026salimory}. Agent Memory Distillation follows a different, training-free strategy. It constructs hierarchical procedural memories from successful trajectories generated by a stronger teacher agent \citep{kim2026amd}.

In these settings, the source experience has already been expressed as text. Long-term egocentric video introduces an additional challenge before memory management can begin. The writer must identify entities, actions, and speech from raw visual and audio observations, preserve their temporal context, and avoid introducing details unsupported by the source. Once evidence is omitted or incorrectly described at this stage, a downstream text-memory architecture cannot recover it without revisiting the video.
Text-input memory systems could therefore complement \textsc{MemLife} by organizing the episodic descriptions after they are written, rather than replacing our system as video-to-text writers.

\section{Theoretical Foundation of FIRM}
\label{app:memory_decomposition}

Let $Q$ and $A$ denote a question and its answer, $V$ the available video history, $M$ the memory written from that history, and $C$ the retrieved context. Let $H(\cdot\mid\cdot)$ and $I(\cdot;\cdot\mid\cdot)$ denote conditional entropy and conditional mutual information. For the memory-only reader, their joint distribution factorizes as
\[
    p(V,Q,A,M,C)=p(V,Q,A)p_\theta(M\mid V)p(C\mid M,Q),
\]
where $\theta$ denotes the writer parameters. This restates Equation~\ref{eq:memory_factorization}. Because the writer constructs $M$ only from $V$ and the reader constructs $C$ only from $M$ and $Q$,
\[
    I(A;M\mid V,Q)=0,
    \qquad
    I(A;C\mid M,Q)=0.
\]
The memory-quality decomposition then follows as
\begin{align}
    & H(A\mid C,Q)-H(A\mid V,Q) \nonumber\\
    ={}& I(A;V\mid Q)-I(A;C\mid Q) \nonumber\\
    ={}& \big[I(A;V\mid Q)-I(A;M\mid Q)\big] +\big[I(A;M\mid Q)-I(A;C\mid Q)\big] \nonumber\\
    ={}& \big[I(A;V\mid M,Q)-I(A;M\mid V,Q)\big] +\big[I(A;M\mid C,Q)-I(A;C\mid M,Q)\big] \nonumber\\
    ={}& I(A;V\mid M,Q)+I(A;M\mid C,Q).
    \label{eq:memory_decomposition_proof}
\end{align}
Applying the chain rule to the two bracketed differences and substituting the conditional independences above yields the final line. Its two terms are the informativeness and retrievability gaps, respectively.

To relate these gaps to downstream QA, let $\hat{A}\sim q_\phi(\cdot\mid Q,C)$ denote the answerer's prediction. The quantity $q_\phi(A\mid Q,C)$ is therefore the probability assigned to the ground-truth answer, while $p_\theta(A\mid Q,C)$ is its conditional distribution under the joint process above. Taking expectations under $p_\theta(Q,A,V,M,C)$, the negative log-likelihood decomposes as
\begin{equation}
\begin{aligned}
\mathcal{L}_{\mathrm{NLL}}(\theta,\phi)
:={}&\mathbb{E}_{p_\theta}\big[-\log q_\phi(A\mid Q,C)\big] \\
={}&\underbrace{H(A\mid Q,V)}_{\text{irreducible uncertainty}}
+\underbrace{I(A;V\mid Q,M)}_{\text{informativeness gap}}
+\underbrace{I(A;M\mid Q,C)}_{\text{retrievability gap}} \\
&+\underbrace{\mathbb{E}_{p_\theta(Q,C)}\,\mathrm{KL}\!\left(
p_\theta(A\mid Q,C)\,\|\,q_\phi(A\mid Q,C)
\right)}_{\substack{\text{prediction mismatch}\\\text{(includes faithfulness effects)}}}.
\end{aligned}
\label{eq:qa_loss_decomposition}
\end{equation}
\noindent\textit{Proof.} For fixed $(Q,C)$, averaging $-\log q_\phi(A\mid Q,C)$ over $A\sim p_\theta(A\mid Q,C)$ gives their cross-entropy. This equals $H(A\mid Q,C)$ plus the KL term in Equation~\ref{eq:qa_loss_decomposition}. Equation~\ref{eq:memory_decomposition_proof} expands the conditional entropy as $H(A\mid Q,V)+I(A;V\mid Q,M)+I(A;M\mid Q,C)$, yielding the result. \hfill$\blacksquare$

The final term is affected by both the memory and the answerer. Unsupported content in $C$ can shift $q_\phi$ away from $p_\theta$ and mislead the answerer, while reasoning errors can produce the same mismatch even when $C$ is fully grounded. It therefore mixes a writer-controlled faithfulness effect with answerer-dependent noise and cannot directly supervise faithfulness. This explains why final-answer feedback gives noisy credit to the writer.

Together, these results motivate the three FIRM signals. Informativeness targets answer-relevant evidence lost during writing, and retrievability targets stored evidence missed during memory access. Faithfulness directly compares the memory with its source, isolating hallucination-related noise without absorbing answerer reasoning error.

For a rigorous assessment of memory quality, \textsc{MemOpt} uses the default \textsc{MemLife} reader during training and disables direct access to the source videos. This prevents the reader from bypassing the written memory when collecting answer-relevant information. 

\section{Human Evaluation of Faithfulness Feedback}
\label{app:rev_human_eval}

The faithfulness evaluator shares its backbone with the writer, so we examine whether its feedback agrees with human judgment. A human annotator inspected 50 spans that the evaluator flagged as unsupported, comparing each span with the frames and transcript of its source segment, and confirmed that 39 of them (78\%) are indeed unsupported. The evaluator can provide this feedback despite the shared backbone because it solves an easier task than the writer. Instead of composing a description, it checks a given description against the source. In addition, training rollouts are sampled at temperature 1.0, so candidates can contain low-probability details that the evaluator, conditioned on the same frames, would not produce. The reward ablation in Table~\ref{tab:reward_ablation} provides complementary evidence, as adding faithfulness to retrievability and informativeness improves accuracy on both benchmarks.

The following example illustrates a correction made by the evaluator. A candidate twice claimed that people holding plates were also drinking from a can, although the source segment supports neither action, and the evaluator removed both claims while keeping the rest of the description.
\begin{quote}\small
\textbf{Before:} ``\ldots another man dressed in black sits cross-legged, holding a plate and drinking from a can. \ldots He also holds a plate and drinks from a can.''

\textbf{After:} ``\ldots another man dressed in black sits cross-legged, holding a plate. \ldots He also holds a plate.''
\end{quote}

Because the evaluator shares the writer's backbone by design, so that \textsc{MemOpt} does not rely on stronger models, errors that this backbone makes systematically remain beyond the reach of the evaluator. In addition, under prefix teacher forcing, the token-level feedback concentrates on the first token of an unsupported span, and span-level credit assignment is a direction for future work.

\section{Qualitative Examples}
\label{app:rev_qualitative}

\subsection{Reader Trajectories}
\label{app:rev_trajectories}

We show representative trajectories of the Qwen3.5-9B reader from the main evaluation, each comparing the \textsc{MemOpt} memory with the zero-shot memory.

\paragraph{(a) One search, then answer.} This is the most common pattern. For the SuperMemory-VQA Conversational Memory question ``What did C say earlier about exactly what sensors the AI glasses use?'', both memories lead to the same first search, which retrieves the evidence segment at a similar rank. The \textsc{MemOpt} entry keeps the speaker labels from the transcript, and the reader answers correctly. The zero-shot entry attributes the statement of C to the wearer, so the reader finds no record of C and answers that the question cannot be answered. Appendix~\ref{app:rev_examples} gives the complete entries.

\paragraph{(b) Query refinement.} For the EgoLifeQA question ``What does Alice not eat?'', with the options noodles, bread, cream, and pizza, a first search with these options returns no entry about what Alice avoids. The reader then rewrites the query around the negation and retrieves an entry in which Alice says, ``I don't eat cream,'' answering correctly in round 3. With the zero-shot memory, five searches never return this entry, and the reader answers incorrectly.

\paragraph{(c) Search, then fetch the surrounding interval.} For a SuperMemory-VQA question asking for the name of the pizza place and the soda taken from the fridge, the first search retrieves two evidence entries, but the name on the pizza box remains unclear. The reader then uses \textsc{FetchMemory} to read the next three minutes in order, finds the box labeled ``Luna Pizza Kitchen'', and answers correctly in round 3.

\paragraph{(d) Failure with retrieved evidence.} For the SuperMemory-VQA Visual Recall question ``I want to check the kitchen specs. What brand was the microwave?'', both memories rank the evidence entry fourth. The zero-shot entry mentions a Frigidaire microwave, and the reader answers correctly. The \textsc{MemOpt} entry describes only a stainless steel microwave, and the reader answers that the question cannot be answered. This example shows that optimization can omit a visual detail needed by a later question, consistent with the decrease on Visual Recall in Table~\ref{tab:rev_category_results}, the category with the fewest training question--segment pairs (Table~\ref{tab:rev_category_supervision}).

\subsection{Complete Memory Entries}
\label{app:rev_examples}

\begin{figure}[h!]
    \centering
    \includegraphics[width=0.32\linewidth]{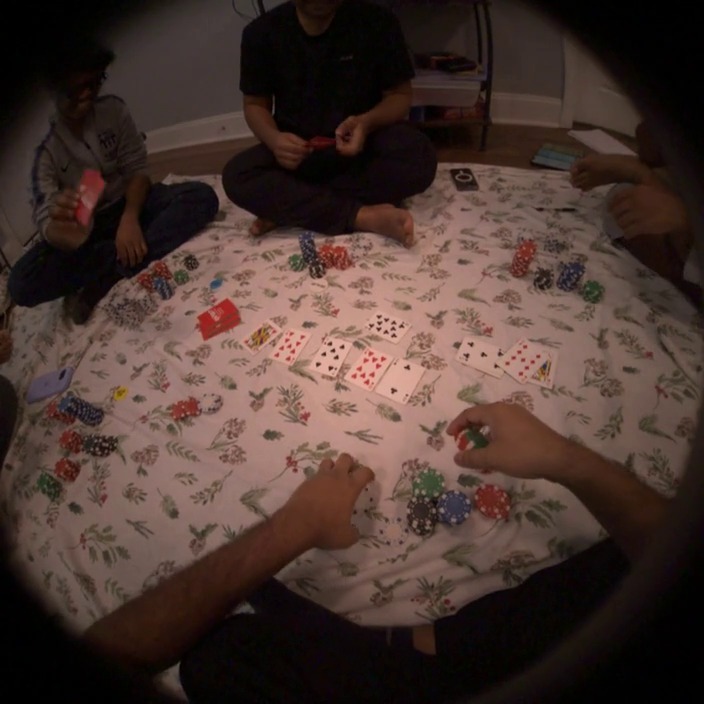}\hfill
    \includegraphics[width=0.32\linewidth]{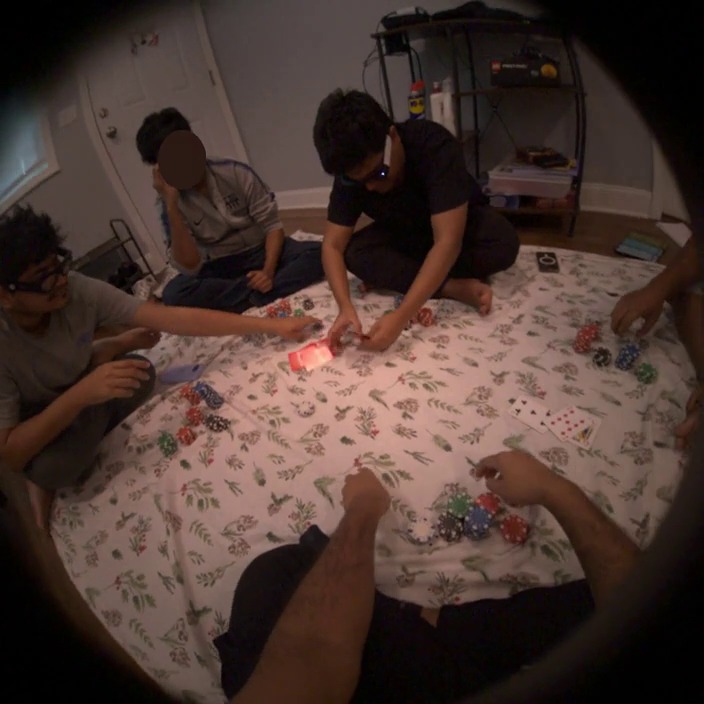}\hfill
    \includegraphics[width=0.32\linewidth]{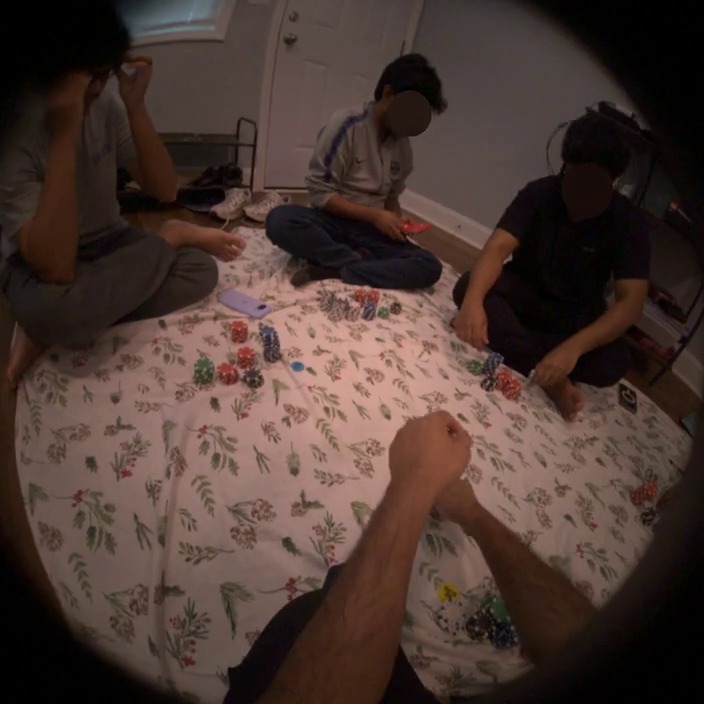}
    \caption{Frames from the evidence segment of example (a) in Appendix~\ref{app:rev_trajectories}.}
    \label{fig:rev_complete_entry_frames}
\end{figure}

We give the complete entries written before and after \textsc{MemOpt} for the evidence segment of example (a), recorded during a card game, together with its transcript. Figure~\ref{fig:rev_complete_entry_frames} shows three of the eight frames given to the writer.

\begin{quote}\small
\textbf{Transcript.} [A] So, like, it is recording the audio and video, right? [C] Audio, video, your eye gaze, and also your head orientation.

\textbf{\textsc{MemLife}.} I am sitting cross-legged on a white sheet with a green and red floral pattern, surrounded by four other people in a room with light blue walls and wooden floors. In front of me are stacks of poker chips in red, blue, green, black, and yellow, along with several playing cards laid out face-up including the 7 of hearts, 10 of spades, and Ace of diamonds. To my left, a person wearing glasses and a grey sweatshirt reaches across to grab a red chip from the center pile while another individual in a dark t-shirt sits opposite me holding a deck of cards. A third person in a navy shirt and light shorts is seated to my right, occasionally moving their hands near their own stack of chips. Behind us stands a metal shelving unit holding various items including what appears to be an electronic device with buttons and dials. The camera captures both the visual scene and audio dialogue where someone asks if the recording includes audio and video, and I respond that it also records eye gaze and head orientation. Throughout this segment, we continue managing our chip stacks and observing the game state as others interact with the central pile of cards and tokens.

\textbf{\textsc{MemLife} + \textsc{MemOpt}.} I am participating in a group activity where we are sitting on a white cloth with a floral pattern that has green leaves and red berries. There are several people around me, including one person wearing a gray long-sleeve shirt with blue stripes on the sleeves and dark pants, another person in a black t-shirt and dark pants who is handling red cards, and others whose faces are obscured. In front of us, there are scattered playing cards face up, showing various suits and numbers such as hearts, spades, and diamonds. There are also multiple piles of poker chips in different colors---red, blue, green, and black---arranged across the cloth. I am actively handling the chips with both hands, picking them up and placing them down. The room has light-colored walls, a white door in the background, and a shelf with items including a bottle, a box, and some electronic equipment. During this segment, [A] asks, ``So, like, it is recording the audio and video, right?'' and [C] responds, ``Audio, video, your eye gaze, and also your head orientation.'' This indicates that the current setup is capturing not only audio and video but also tracking my eye movements and head position. The overall atmosphere appears casual and focused on the game or activity we are engaged in, with everyone involved in handling cards or chips.
\end{quote}

\section{Prompt Templates}
\label{app:prompt_and_tool_details}

\subsection{Writer Prompt}
\label{app:writer_prompt}

Figure~\ref{fig:prompt_writer} shows the complete system instruction and user template used by the \textsc{MemLife} writer. 
The prompt asks it to preserve visible and spoken evidence, ground named entities in visual observations, and narrate the resulting memory in the first person.

\begin{figure}[h!]
\begin{center}
\begin{AIbox}{Prompt for the \textsc{MemLife} memory writer}
\raggedright
\textit{System instruction}\\

You are writing a memory entry for one short segment of an egocentric video. This entry is the only record of this moment that a later question-answering system will see, so capture the concrete, factual details it might be asked about.\\

Write in the FIRST PERSON as the person wearing the camera (the video is shot from their point of view): refer to yourself as `I'/`my'/`me', not as `the camera wearer', `the camera operator', or `the person'.\\

The evidence lives in TWO equally important channels, and you must use both:\\
1. What is SEEN --- the people, their actions, the objects they handle (brand names, text on labels, colors), and the place. Refer to any OTHER person by name when they are a named speaker in the transcript (tags like [B] or [Shure]); add stable visual attributes (clothing, hair, glasses) as a complement to the name, and use visual attributes alone only when no name is available.\\
2. What is SAID or HEARD --- who speaks and the substance of what they say; paraphrase salient statements, questions, decisions, plans, names, numbers, times, and places mentioned in the transcript.\\

Treat the spoken transcript as a primary source of fact, not background noise: if something is only stated aloud, it still belongs in the memory. Write ONE paragraph of plain prose, up to $\sim$450 tokens; finish every sentence and do not truncate. No JSON, no bullets, no headers. Describe only this segment; do not speculate beyond the evidence.\\

Write a COMPLETE, SELF-CONTAINED first-person paragraph describing what happens in this segment (actions, objects, places, people, state changes).\\[0.5em]

\textit{User template}\\

\verb|[eight frames sampled from the segment, 704 px]|\\

What is said in this segment:\\
\verb|"""{transcript}"""|\\

Write the memory entry, grounding it in the frames and transcript above.
\end{AIbox}
\end{center}
\caption{Complete system instruction and user template for the \textsc{MemLife} memory writer.}
\label{fig:prompt_writer}
\end{figure}

\subsection{Reader System Prompts}
\label{app:reader_prompts}

Figures~\ref{fig:prompt_reader_agent} and \ref{fig:prompt_reader_video} provide the agentic-reader prompts used in our experiments.
The reader emits either one tool call or a final answer per round. Outputs that match neither format are treated as parse failures and scored as incorrect. All requested intervals are clipped to the question time, preventing access to future segments.

\begin{figure}[h!]
\begin{center}
\begin{AIbox}{System prompt for the default \textsc{MemLife} reader}
\raggedright
You are a memory reader. You answer multiple-choice questions about an egocentric video by calling tools to inspect a pre-built memory.\\

You have two tools:\\

1. \verb|search_memory(query: str,|\\
\verb|   time_anchor: [t_start, t_end] = [0, <now>],|\\
\verb|   top_k: int = 32)|:\\
retrieve the memory entries most relevant to the query, restricted to clips within \verb|[t_start, t_end]|. Returns the \verb|top_k| most similar clips in time order --- each with its description and time range. Omit \verb|time_anchor| to search all memory up to the current time; pass a narrower \verb|[t_start, t_end]| to focus on a period (e.g. a specific day).\\

2. \verb|fetch_memory(|\\
\verb|   time_anchor: [t_start, t_end])|:\\
return every memory clip within \verb|[t_start, t_end]| in time order, each with its description.\\

Each turn, respond with EXACTLY ONE of these two lines:\\
\verb|  TOOL: {"name": "<tool>", "args": {...}}|\\
\verb|  ANSWER: <letter>|\\

Emit no other text on the response line. End your final round with \verb|`ANSWER: <letter>`| once you have enough evidence.\\

Examples:\\
\verb|Q: "How often do I cook dinner?" ->|\\
\verb|TOOL: {"name": "search_memory",|\\
\verb|       "args": {"query": "cooking dinner"}}|\\
\verb|Q: "What did I buy yesterday?" ->|\\
\verb|TOOL: {"name": "search_memory",|\\
\verb|       "args": {"query": "buying purchase shopping",|\\
\verb|                "time_anchor": [yesterday_start,|\\
\verb|                                yesterday_end]}}|\\
\verb|Q: "List everything I bought today." ->|\\
\verb|TOOL: {"name": "fetch_memory",|\\
\verb|       "args": {"time_anchor": [t_start, t_end]}}|
\end{AIbox}
\end{center}
\caption{System prompt for the default \textsc{MemLife} reader.}
\label{fig:prompt_reader_agent}
\end{figure}

\begin{figure}[h!]
\begin{center}
\begin{AIbox}{Additional source-video tool for \textsc{MemLife}-V}
\raggedright
3. \verb|fetch_video(|\\
\verb|   time_anchor: [t_start, t_end],|\\
\verb|   num_frames: int = 30)|:\\
sample \verb|num_frames| frames and the transcript from the video within \verb|[t_start, t_end]|.\\

\verb|Q: "What was on my desk when I left?" ->|\\
\verb|TOOL: {"name": "fetch_video",|\\
\verb|       "args": {"time_anchor": [t_start, t_end],|\\
\verb|                "num_frames": 30}}|
\end{AIbox}
\end{center}
\caption{Additional prompt block for \textsc{MemLife}-V. The tool count is updated to three, and all other text follows Figure~\ref{fig:prompt_reader_agent}.}
\label{fig:prompt_reader_video}
\end{figure}

\subsection{Evaluator Prompts}
\label{app:evaluator_prompt_templates}

Figures~\ref{fig:prompt_key_fact}--\ref{fig:prompt_faithfulness} provide the exact templates for grounded key-fact extraction, informativeness evaluation, and faithfulness evaluation. The key-fact extractor and faithfulness evaluator receive eight frames and the transcript of the current 30-second segment, whereas the entailment judge receives only the key fact and candidate memory.

\begin{figure}[h!]
\begin{center}
\begin{AIbox}{Prompt template for grounded key-fact extraction}
\textit{System instruction}\

You are auditing ONE 30-second segment of a first-person video against a question about the wearer's day and that question's known-correct answer. The segment has been labelled as evidence for the question. State, in one sentence, the KEY FACT that is actually visible or audible in THIS segment and that helps establish the correct answer.\\

Rules:\\
1. The key fact must be something you can SEE in this segment's frames or HEAR in its transcript. Never infer it from the question or from the answer.\\
2. Write it as a standalone declarative sentence about the segment, naming the concrete objects, people, places, actions or spoken words involved. Do NOT mention the question, the options, or the answer, and do not reuse an option's wording.\\
3. If this segment does NOT actually contain anything supporting the correct answer - the evidence label is wrong, or the segment is unrelated background - say so instead of inventing a link.\\
4. A segment can be genuine evidence while showing only PART of what the answer needs; in that case state the part it does show.\\

Output EXACTLY two lines and nothing else:\\
SUPPORTS: YES\\
KEY FACT: \verb|<one sentence>|\\
or\\
SUPPORTS: NO\\
KEY FACT: NONE\\[0.5em]

\textit{User template}\

\#\# This segment\\
\verb|{eight frames sampled from the segment}|\\

What is said in this segment:\\
\verb|"""{transcript}"""|\\

\#\# Question a later reader will ask about this day\\
\verb|"{question}"|\\
Options:\\
\verb|{options}|\\
Correct answer: \verb|{letter}. {correct_answer}|\\

What key fact does THIS segment contribute toward that correct answer?\\
Output the two lines.
\end{AIbox}
\end{center}
\caption{Prompt template for grounded key-fact extraction.}
\label{fig:prompt_key_fact}
\end{figure}

\begin{figure}[h!]
\begin{center}
\begin{AIbox}{Prompt template for informativeness evaluation}
\textit{System instruction}\

You are a strict entailment judge. You are given a TARGET FACT about one 30-second segment of a first-person video, and a MEMORY that was written for that same segment. Decide whether the memory states the target fact, or states something from which the target fact follows directly. A paraphrase counts. A vaguer statement that omits the specific detail in the target fact does not. Use only these two texts - no outside knowledge and no guessing. Answer with exactly one word: Yes or No.\\[0.5em]

\textit{User template}\

\#\# TARGET FACT\\
\verb|"{key_fact}"|\\

\#\# MEMORY\\
\verb|"{candidate_memory}"|\\

Does the memory state or directly entail the target fact? Answer Yes or No.\\
Answer:
\end{AIbox}
\end{center}
\caption{Prompt template for informativeness evaluation.}
\label{fig:prompt_sufficiency}
\end{figure}

\begin{figure}[h!]
\begin{center}
\begin{AIbox}{Prompt template for faithfulness evaluation}
\textit{System instruction}\

You are correcting a written memory of one 30-second segment of a first-person video. You are given the segment's frames and transcript, and ONE memory of that segment. A claim is unsupported if it asserts an object, person, place, action, quantity, or spoken content that is not actually visible in the segment's frames or audible in its transcript. A memory that is vague, brief or incomplete is still supported.\\

If every claim in the memory is supported, repeat the memory back EXACTLY as given, character for character. If some claim is not supported, output the memory with those claims removed, or weakened to what the segment actually shows, changing as few words as possible. Do not add detail and do not improve the writing.\\

Output only the memory text. No preamble, no explanation, no quotation marks.\\[0.5em]

\textit{User template}\

\#\# The segment\\
\verb|{eight frames sampled from the segment}|\\

What is said in this segment:\\
\verb|"""{transcript}"""|\\

\#\# Memory of this segment\\
\verb|"{candidate_memory}"|\\

Output this memory with any unsupported claims corrected, or unchanged if every claim is supported.\\
Memory:
\end{AIbox}
\end{center}
\caption{Prompt template for faithfulness evaluation.}
\label{fig:prompt_faithfulness}
\end{figure}

\end{document}